\documentclass[10pt,twocolumn,letterpaper]{article}

\usepackage[pagenumbers]{cvpr} % To force page numbers, e.g. for an arXiv version

\usepackage[dvipsnames]{xcolor}
\usepackage{multirow}
\usepackage{booktabs}
\usepackage{cellspace}
\usepackage{algorithm}
\usepackage{algorithmic}
\usepackage{bbding}
\usepackage{amssymb}
\usepackage{graphicx}
\usepackage{caption,subcaption}
\usepackage{colortbl}
\usepackage[accsupp]{axessibility}

\newcommand{\cls}[1]{{\small\texttt{#1}}}

\newcommand{\bluecomment}[1]{\textcolor{blue}{$\triangleright$ #1}}

\definecolor{cvprblue}{rgb}{0.21,0.49,0.74}
\usepackage[pagebackref,breaklinks,colorlinks,citecolor=cvprblue]{hyperref}

\def\paperID{6252} % *** Enter the Paper ID here
\def\confName{CVPR}
\def\confYear{2024}

\title{PTQ4SAM: Post-Training Quantization for Segment Anything}

\author{Chengtao~Lv\textsuperscript{1}$^{*}$, Hong~Chen\textsuperscript{1}$^{*}$, 
Jinyang~Guo\textsuperscript{1,3}$^{\dagger}$,
Yifu~Ding\textsuperscript{1},
Xianglong~Liu\textsuperscript{1,2,4}\\
\textsuperscript{1} State Key Laboratory of Complex \& Critical Software Environment, Beihang University \\
\textsuperscript{2} Zhongguancun Laboratory \quad \textsuperscript{3} Institute of Artificial Intelligence, Beihang University \\
\textsuperscript{4} Institute of data space, Hefei Comprehensive National Science Center \\
}

\begin{document}
\maketitle
\newcommand\blfootnote[1]{% 
\begingroup 
\renewcommand\thefootnote{}\footnotetext{#1}% 
\addtocounter{footnote}{0}% 
\endgroup 
}
\blfootnote{$^*$ Equal contribution.}
\blfootnote{$^{\dagger}$ Corresponding author: jinyangguo@buaa.edu.cn}
\begin{abstract}
Segment Anything Model (SAM) has achieved impressive performance in many computer vision tasks. However, as a large-scale model, the immense memory and computation costs hinder its practical deployment. In this paper, we propose a post-training quantization (PTQ) framework for Segment Anything Model, namely PTQ4SAM. First, we investigate the inherent bottleneck of SAM quantization attributed to the bimodal distribution in \cls{post-Key-Linear} activations. We analyze its characteristics from both per-tensor and per-channel perspectives, and propose a Bimodal Integration strategy, which utilizes a mathematically equivalent sign operation to transform the bimodal distribution into a relatively easy-quantized normal distribution offline. Second, SAM encompasses diverse attention mechanisms (i.e., self-attention and two-way cross-attention), resulting in substantial variations in the post-Softmax distributions. Therefore, we introduce an Adaptive Granularity Quantization for Softmax through searching the optimal power-of-two base, which is hardware-friendly. Extensive experimental results across various vision tasks (instance segmentation, semantic segmentation and object detection), datasets and model variants show the superiority of PTQ4SAM. For example, when quantizing SAM-L to 6-bit, we achieve lossless accuracy for instance segmentation, about 0.5\% drop with theoretical 3.9$\times$ acceleration. The code is available at \url{https://github.com/chengtao-lv/PTQ4SAM}.
\end{abstract}
% \jy{Distribution}
% intro 解释post-Key-Linear
% 加逗号，写成2 perspective
% We discover the bottlenecks of quantization mainly lie in the Bimodal Distribution in SAM. The distribution occur in 
% we investigate the inherent bottleneck of SAM quantization attributed to the bimodal distribution in \cls{post-Key-Linear}.
%Second, the diversity in attention mechanisms
% histogram-based discriminator to automatically detect bimodal distributions
% scheme through searching the optimal hardware-friendly base, which is 
%through dynamically searching the optimal hardware-friendly base
% We refer to it as \textit{Bimodal  Distribution} and have applied a simple $\operatorname{sign}$ operation to transform it into a easy-quantized \textit{Single Peak Distribution}. Second,... Extensive experimental results across various vision tasks (instance segmentation, semantic segmentation, object detection and tracking) and models show the superiority of PTQ4SAM.     
\section{Introduction}
\label{sec:intro}

With remarkable zero-shot ability and user-friendly flexible prompt technique, Segment Anything Model (SAM) \cite{kirillov2023segment} has recently become a novel foundation model in a range of generic vision applications, including image segmentation \cite{zhang2023personalize,cheng2023sam,lan2023foodsam,li2023semantic}, object detection \cite{wang2023detect,tang2023can}, tracking \cite{maalouf2023follow,yang2023track,cheng2023segment} and other downstream tasks \cite{yu2023inpaint,shen2023anything,yao2023matte,wang2021dual,liu2019perceptual,liu2022harnessing}. However, the transformer architectures in SAM require intensive computation and memory footprint, which hurdles the practical deployment on resource-constrained edge-devices.
% However, the transformer architectures in SAM, which require intensive computation and memory footprint, will surpass 2GB in size and the computational complexity may reach the order of TFLOPs. This poses a challenge for the practical deployment of SAM on edge-devices with limited resources.
%However, the transformer architectures in SAM require intensive computation and memory footprint, which hurdles the practical deployment on  resource-constrained edge-devices.

% In this paper, we study quantization to compress the neural networks which converts weights and activations from floating-point to low-bit \cite{gholami2022survey,nagel2021white}.
To address this issue, several quantization approaches~\cite{gholami2022survey,nagel2021white,choukroun2019low,li2023repq,li2021brecq,nagel2020up,jacob2018quantization,wei2022qdrop,wu2020integer} were proposed to convert weights and activations from floating-point to low-bit. There are two categories of quantization methods: 1) Quantization-Aware Training (QAT) and 2) Post-Training Quantization (PTQ). QAT retrains a model by utilizing the whole labeled training dataset, which will be time-consuming due to the corresponding massive dataset (SA-1B). On the other hand, PTQ is more promising because it only requires small unlabeled samples to calibrate the pre-trained networks. In this paper, we focus on designing the PTQ approach as it is more effective in practical usage.

\begin{figure}[t]
\centering
\subcaptionbox{bimodal distribution \label{fig:intro_a}}{\includegraphics[width=.48\linewidth]{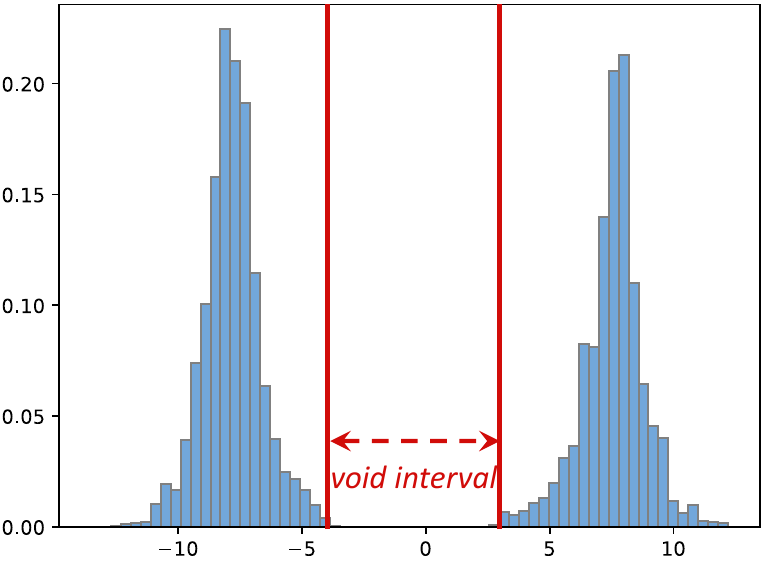}}
\subcaptionbox{post-Softmax distribution \label{fig:intro_b}}{\includegraphics[width=.48\linewidth]{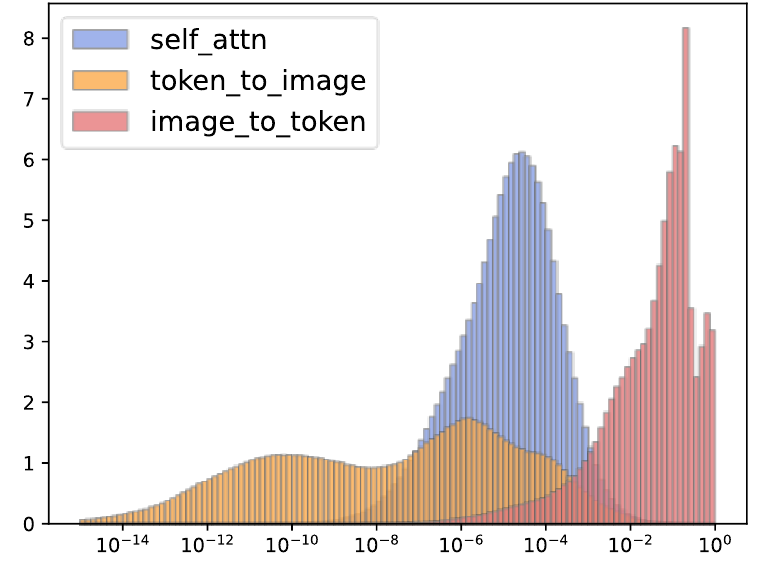}}

% \end{center}
 % \captionsetup{skip=0pt}
 % \captionsetup{belowskip=0pt}
  \caption{The histogram of two special distributions in SAM: (a) bimodal distribution in \cls{post-Key-Linear} activations.  (b) post-Softmax distributions of self-attention, token-to-image cross-attention and image-to-token cross-attention.}
  % Grey lines are quantization levels.
  % \vspace{-1mm}
 %and cross-attention in two directions
    \label{fig:intro}
\end{figure}

Although previous PTQ methods have showcased significant accomplishments in various scenarios, including convolutional neural networks (CNNs) \cite{nagel2020up,li2021brecq,wei2022qdrop,wu2020integer,choukroun2019low}, vision transformers (ViTs) \cite{yuan2022ptq4vit,ding2022towards,li2023repq} as well as large language models (LLMs) \cite{xiao2023smoothquant,wei2022outlier,wei2023outlier}, directly adopting these methods to Segment Anything Models will raise two unique challenges that necessitate a revisiting of traditional PTQ schemes: 1) we observe the bimodal distribution appears in \cls{post-Key-Linear} activations, \ie, the output activations of key linear, as shown in Figure \ref{fig:intro_a}. The two peaks and their central void interval severely enlarge the range of the entire distribution, which negatively affects quantization performance. 2) Owing to the diverse kinds of attention mechanisms, SAMs exhibit more complicated post-Softmax distributions when compared to VITs, as shown in Figure \ref{fig:intro_b}. Statistically, about 72.5\% post-Softmax activations of \cls{image-to-token} are more than 0.01 while only 0.4\% in \cls{token-to-image}. Prior works \cite{ding2022towards,yuan2022ptq4vit,li2023repq,lin2021fq} have not adequately addressed this discrepancy and treated them equally, causing the potential loss of inherent information. Therefore, it is desirable to design specialized components for post-Softmax distribution in SAM.
% If we do not consider this discrepancy and use the same quantizer to quantize different post-Softmax activations like previous works~\cite{ding2022towards,li2023repq}, it will inevitably cause inherent information loss. 
% Prior works have not adequately addressed this discrepancy and treat them equally, causing the potential loss of inherent information. Therefore, the specialized components for post-Softmax distribution in SAM is desirable.

% 1.causing 2.therefore 重新设计
% finding that cruial 
% , based on which we utilize 
% 把判定揉到创新点里面
% fp最左边 加个gt

% \jy{the bimodal distribution exists in a portion of tensors, which is the crucial obstacle of SAM quantization.} 
% \jy{We leverage its per-tensor characteristic to determine whether a distribution bimodal distribution, }
% \jy{and use its per-channel characteristic to transfer these bimodal distribution to a normal distribution that is easy to quantized \textit{offline}.}
% based on which the channel-wise sign factor is simultaneously absorbed into the query linear and key linear \textit{offline}.
% \jy{This transformation is mathematically equivalent and can significantly narrowing the distribution range for better quantization performance.}
% The operation is mathematically equivalent and transforms the bimodal distribution to normal distribution, significantly narrowing the distribution range. 
% \jy{To address the second issue,} 

Based on the above observations, in this paper, we propose a novel post-training framework called PTQ4SAM specifically designed for Segment Anything Model quantization. First, we present a \textit{Bimodal Integration} (BIG) strategy to deftly eliminate the bimodal distribution. Specifically, we conduct an in-depth analysis of bimodal distribution from both per-tensor and per-channel perspectives, finding that the bimodal distribution is the crucial obstacle to SAM quantization. We leverage its per-tensor characteristic to determine whether a distribution bimodal distribution, and use its per-channel characteristic to transfer this bimodal distribution to a normal distribution by simultaneously absorbing sign factor into query linear and key linear \textit{offline}. This transformation is mathematically equivalent and significantly narrows the distribution range for better quantization performance. Second, we propose an \textit{Adaptive Granularity Quantization} (AGQ) explicitly tailored for diverse post-Softmax distributions, which affords a suitable trade-off in granularity for both lower and higher attention scores. We provide theoretical proof for its efficiency on hardware by searching the optimal power-of-two base. Instead of minimizing quantization errors of attention scores, we design the objective by the matrix multiplication output between attention scores and values, which is more robust and beneficial to the ultimate performance. 

We conduct extensive experiments on fundamental tasks and different model variants to demonstrate the versatility of PTQ4SAM. Our PTQ4SAM can seamlessly plug into both statistic-based and learning-based PTQ methods, achieving 3.9× FLOPs and 4.9× storage savings while maintaining lossless performance on 6-bit SAM-L and SAM-H. Our major contributions are summarized as follows:
% , achieving satisfactory performance 
\begin{itemize}
    \item To our best knowledge, our work is the first post-training quantization solution tailored for Segment Anything Model, dubbed PTQ4SAM. % fully不准确 位置不合适
    \item We observe a challenging bimodal distribution for quantization and analyze its characteristics. To overcome it, we propose a Bimodal Integration (BIG) strategy, which automatically detects it and transforms the bimodal distribution to normal distribution equivalently.
    % \item We observe a challenging bimodal distribution and analyze its characteristic, finding that this bimodal distribution is the critical obstacle for SAM quantization. To alleviate this, we propose a Bimodal Integration (BIG) strategy, which automatically detects it and transforms the bimodal distribution to normal distribution equivalently. % which 很难量化 challenge 太细了 写成转成单峰 判别器occurrence location
    \item We present the Adaptive Granularity Quantization (AGQ) which represents diverse post-Softmax distributions accurately with appropriate granularity. %  精确地 搜索 硬件友好 
    % \item Comprehensive experiments conducted on various tasks, all SAM variants and challenging bit-widths. Our PTQ4SAM is a plug-and-play method and significantly surpasses previous state-of-art quantization schemes by a large margin.
    \item Comprehensive experiments conducted on various tasks, variants, and bit-widths demonstrate our PTQ4SAM is a plug-and-play method and significantly surpasses previous state-of-the-art PTQ schemes by a large margin.
\end{itemize}

\section{Related Work}
\label{sec:relatedwork}

% All text must be in a two-column format.
% The total allowable size of the text area is $6\frac78$ inches (17.46 cm) wide by $8\frac78$ inches (22.54 cm) high.
% Columns are to be $3\frac14$ inches (8.25 cm) wide, with a $\frac{5}{16}$ inch (0.8 cm) space between them.
% The main title (on the first page) should begin 1 inch (2.54 cm) from the top edge of the page.
% The second and following pages should begin 1 inch (2.54 cm) from the top edge.
% On all pages, the bottom margin should be $1\frac{1}{8}$ inches (2.86 cm) from the bottom edge of the page for $8.5 \times 11$-inch paper;
% for A4 paper, approximately $1\frac{5}{8}$ inches (4.13 cm) from the bottom edge of the
% page.

%-------------------------------------------------------------------------
\subsection{Segment Anything}
Recently, Meta AI Research has revolutionarily approached a general, promptable Segment Anything Model (SAM) \cite{kirillov2023segment}. Pre-training on web-scale datasets (SA-1B), SAM demonstrates the capability to generalize across diverse downstream tasks \cite{cheng2023sam,yu2023inpaint,shen2023anything,yang2023track,wang2023caption}. HQ-SAM \cite{ke2023segment} designs learnable tokens and global-local fusion schemes to obtain high-quality masks. SEEM \cite{zou2023segment} extends the referring image to prompt types and integrates a joint visual-semantic space. In the realm of medical research, MedSAM \cite{ma2023segment} and SAM-Med2D \cite{cheng2023sam} fine-tune SAM through large-scale medical image datasets. Combined with a series of visual-language models \cite{mildenhall2021nerf,li2022blip,rombach2022high}, Anything-3D \cite{shen2023anything} and SA3D \cite{cen2023segment} applies SAM to the single-view 3D reconstruction task while Seal \cite{liu2023segment} to the 3D point cloud segmentation task. Suffering from substantial computational requirements, some efficient SAMs, including MobileSAM~\cite{zhang2023faster}, FastSAM~\cite{zhao2023fast} and TinySAM~\cite{shu2023tinysam} are successively introduced. However, SAM still undergoes untenable resource-intensive consumption. Its real-time processing capabilities have received widespread expectations.
% Moreover, these approaches aims to design different network architecture for different tasks or goals. Our PTQ4SAM aims to quantize an existing SAM architecture for efficient inference, which can be potentially combined with them.

% Moreover, these approaches aims to design different network architecture for different tasks or goals. Our PTQ4SAM aims to quantize an existing SAM architecture for efficient inference, which can be potentially combined with them.
% However, SAM still undergoes untenable resource-intensive consumption. Its real-time processing capabilities have drawn widespread expectation.
%-------------------------------------------------------------------------
\subsection{Post-Training Quantization}
% Quantization refers to using fewer bits to represent 32-bit full-precision floating point numbers, thereby compressing the model size and reducing calculation consumption. 
% Compared with the QAT method, post-training quantization (PTQ) is a simple and efficient quantization method that does not require complex training and large amounts of data. 

% Depending on whether the parameters of the original model are fine-tuned
As a predominant compression approach~\cite{guo2020multi,guo2020channel,guo2023cbanet,guo2021jointpruning,guo2020model,guo2023multidimensional}, the mainstream post-training quantization (PTQ) methods can be broadly divided into two categories~\cite{zheng2022leveraging, niu2023improving}: statistic-based PTQ and learning-based PTQ. Statistic-based PTQ methods solely seek optimal quantization parameters to minimize quantization errors, whereas learning-based PTQ methods fine-tune both weights and quantization parameters. Our approach is out-of-the-box on both kinds of methods.
% while simultaneously exploring for optimal quantization parameters. 
% Learning-based PTQ methods only add a little training overhead in the quantization process, but tend to achieve better results. 
% Our approach is easy to implement, plug-and-play and applicable to both categories of methods.
% \jy{Our approach is applicable to both categories of methods.}
% For CNNs, a lot of works~\cite{jacob2018quantization, mckinstry2019discovering, choukroun2019low} have demonstrated that quantization methods have efficient inference and minimal loss in precision.
% Recent studies on LLM quantization have attracted widespread attention. Some PTQ methods for LLMs only quantzing weights~\cite{lin2023awq,gptq,llmint8}, which may also consider the distribution of activations~\cite{lin2023awq} or outliers~\cite{llmint8}.
% Many works choose to quantizing both the weights and activations of LLMs~\cite{zeroquant,xiao2023smoothquant,wei2023outlier,shao2023omniquant}. For instance, 
% OS~\cite{wei2022outlier} and OS+~\cite{wei2023outlier} perform equivalent transformation for the outlier value with the help of layernorm in the language model, making the activation more suitable for quantization. 
\subsubsection{Statistic-Based PTQ} 

Numerous classic statistic-based quantization methods~\cite{jacob2018quantization, mckinstry2019discovering, choukroun2019low,nagel2021white,miyashita2016convolutional} have been shown to achieve minimal loss in precision primarily for convolutional neural networks. However, with the widespread popularity of networks featuring novel architectures, researchers have introduced quantization schemes specifically designed for these networks. When quantizing ViTs, twin uniform quantization~\cite{yuan2022ptq4vit}, Log-Int-Softmax~\cite{lin2021fq}, scale reparameterization~\cite{li2023repq} and matthew-effect preserving quantization~\cite{ding2022towards} are proposed to tackle the output distribution from softmax. The LLM quantization techniques include weight-only quantization~\cite{lin2023awq,gptq,llmint8,chen2024db}, weight and activation quantization~\cite{zeroquant,xiao2023smoothquant,wei2023outlier,wei2022outlier}, aiming to settle the outlier issue from the activations. PTQ4DM~\cite{shang2023post}, Q-Diffusion~\cite{li2023q} discover the variations in the activations during multiple denoising steps in Diffusion and design specialized calibration strategies. However, there is a notable gap between these distinctive distributions and SAMs, and we are the first to explore quantizing SAMs for efficient inference.

\subsubsection{Learning-Based PTQ}
Based on statistic-based PTQ methods, several learning-based PTQ schemes were also proposed. AdaRound~\cite{nagel2020up} optimizes the rounding operation when quantizing weights to minimize the overall loss of the model. Subsequently, many methods are proposed based on AdaRound.  BRECQ~\cite{li2021brecq} proposes a block-wise reconstruction algorithm to optimize the quantized model. QDrop~\cite{wei2022qdrop} introduces drop operation during the reconstruction process to increase the flatness of the optimized model. PD-Quant~\cite{liu2023pd} introduces global information when optimizing quantization parameters. MRECG~\cite{ma2023solving} focuses on the oscillation problem in PTQ and FlexRound~\cite{lee2023flexround} proposes a new learnable weight rounding scheme on large language models for the first time. Unfortunately, these techniques are mainly carried out based on CNN architecture models. Transformer architecture models like SAM remain unexplored.
% The joint optimization of modules considering module capacity leads to improved performance. 
% Vits such as SAM remain unexplored. Researches related to learning-based PTQ are mainly carried out based on CNN architecture models.

% However, these Learning-Based PTQ studies are all about models based on CNN architecture. SAM based on transformer architecture remain unexplored.
\section{Method}

In this section, we will introduce our PTQ4SAM in detail. First, we introduce uniform quantization and logarithmic quantization in Section \ref{sec:preliminaries}. Subsequently, we present an \textit{Bimodal Integration} (BIG) strategy to eliminate bimodal distribution in Section \ref{sec:sign}. Finally, we analyze the discrepancy of diverse post-softmax distribution and propose \textit{Adaptive Granularity Quantization} (AGQ) in Section \ref{sec:log}.

\subsection{Preliminaries}
\label{sec:preliminaries}

\textbf{Basic Notations.}
We use $\boldsymbol{X}$ to represent a matrix, whereas the vectors are marked by $\boldsymbol{x}$. The operator $\odot$ is used to represent element-wise multiplication between matrices or vectors and operator $\cdot$ denotes scalar multiplication. Also, we use $\boldsymbol{XW}$ to denote matrix multiplication. 
%Also, we use $\textbf{\textit{XW}}$ and $\textbf{\textit{Wx}}$ to respectively denote matrix multiplication and matrix-vector multiplication.

\noindent
\textbf{Post-training Quantization.}
Post-training quantization is a prevalent approach to compress the pre-trained neural network. In this paper, we merely study the hardware-efficient quantization methods. For uniform quantization, quantization and de-quantization operations can be defined as:
% \begin{equation}
%     x_{q} = \mathtt{clamp}( \lfloor \frac{x}{s} \rceil+z , 0, 2^k-1),
% \end{equation}
% \begin{equation}
%     \hat{x} = s \cdot (x_{q} - z) \approx x,
% \end{equation}
\begin{gather}
    x_{q} = \mathtt{clamp}( \lfloor \frac{x}{s} \rceil+z , 0, 2^k-1),\\
    \hat{x} = s \cdot (x_{q} - z) \approx x,
\end{gather}
where $s$ and $z$ denote the scaling factor and zero point, respectively. $\lfloor \cdot \rceil$ is the round-to-nearest operator. $x$ and $\hat{x}$ are floating-point and de-quantized values, and $x_q$ is mapped integer.  $\mathtt{clamp}$ function clips the values fall outside the range of a $k$-bit integer. 

In light of rapid bit-shifting operations, Log2 Quantization has emerged as an alternative hardware-oriented quantization approach. Due to the Log2 Quantization is exclusively employed on post-Softmax activations, it is simply formulated as:
\begin{gather}
    x_{q} = \mathtt{clamp}( \lfloor - \log_2 \frac{x}{s} \rceil, 0, 2^k-1),\\
    \hat{x} = s \cdot 2^{-x_{q}} \approx x.
\end{gather}
% \begin{equation}
%     x_{q} = \mathtt{clamp}( \lfloor - \log_2 \frac{x}{s} \rceil, 0, 2^k-1),
% \end{equation}

% \begin{equation}
%     \hat{x} = s \cdot 2^{-x_{q}} \approx x.
% \end{equation}

% , consisting of uniform quantization and logarithmic quantization.  \tilde{x}

% SAM is also an encoder-decoder architecture based on the attention mechanism. Its computational load is mainly concentrated on matrix multiplication in the fully connected layer and attention function.
\begin{figure*}[]
    \centering
    \includegraphics[width=1\linewidth]{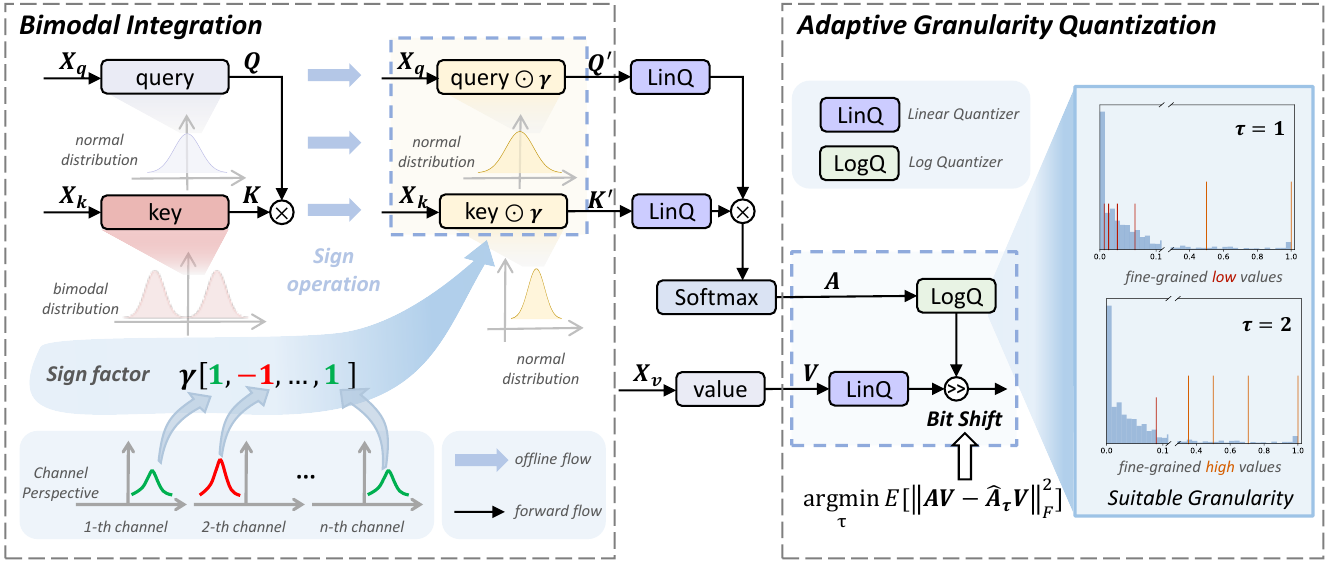}
    \caption{Illustration of our proposed PTQ4SAM. The Bimodal Integration eliminates the bimodal distribution by simultaneously multiplying a channel-wise $\boldsymbol{\gamma}$ to both the query and key linears. The Adaptive Granularity Quantization is employed for post-softmax distribution.}
    \label{fig:frame}
    \vspace{-1mm}
\end{figure*}

\subsection{Bimodal Integration} % kernel mergence/fusion
\label{sec:sign}

\begin{figure}[t]
\begin{center}
     \includegraphics[width=0.9\linewidth]{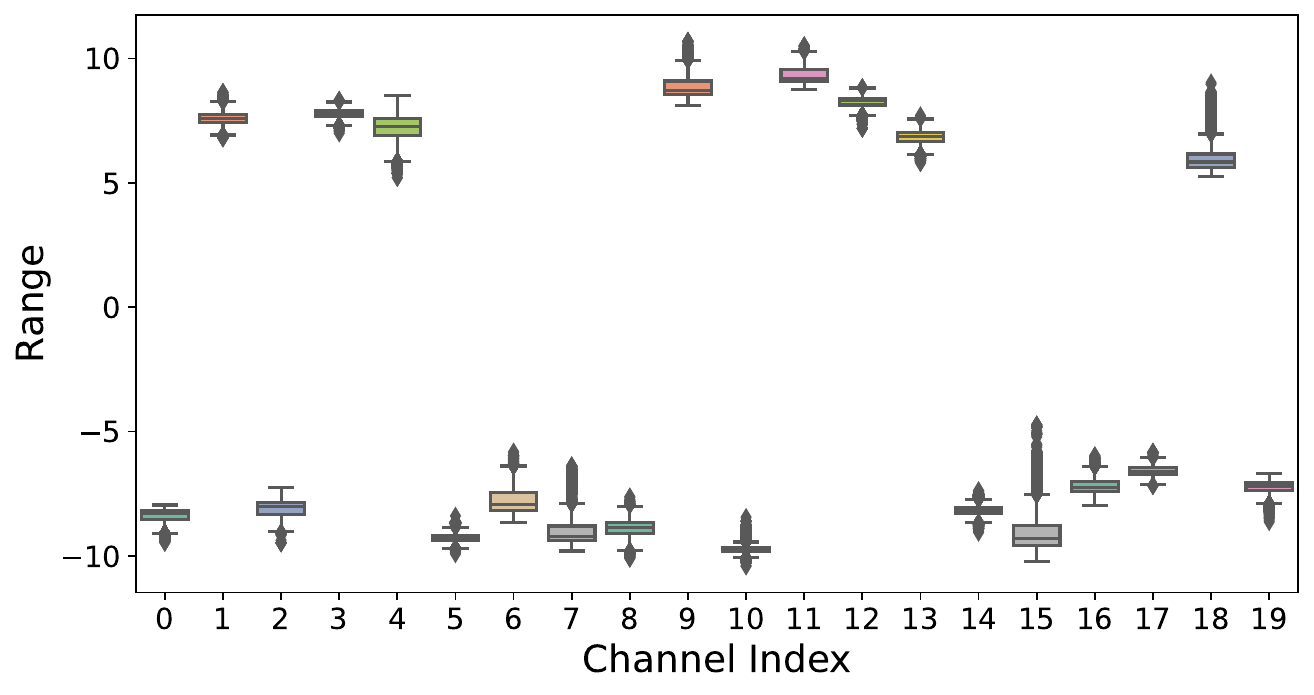}
\end{center}
 \captionsetup{skip=0pt}
  \caption{Boxplot of different channels of \cls{post-Key-Linear} activations in SAM. }
    \label{fig:box}
    \vspace{-1mm}
\end{figure}

Intuitively, the bimodal distribution poses significant challenges for quantization. The two peaks, accompanied by their central void or sparse interval considerably expand the distribution range, leading to over 5$\times$ quantization errors compared to normal distribution experimentally. Therefore, we first make an in-depth analysis of bimodal distribution in SAM from two perspectives: 1) From \textbf{per-tensor} perspective: the distribution contains two peaks and their centers are symmetric, \eg, -8 and 8 in Figure \ref{fig:intro_a}. 2) From \textbf{per-channel} perspective: the activations of each channel only persist in a fixed peak, indicating pronounced asymmetry inside one channel. As shown in Figure \ref{fig:box}, for instance, the activations of the $0$-th channel correspond to a negative peak (\ie, -8), while the activations of $1$-th channel belong to a positive peak (\ie, 8). Generally, about half channels (\eg, 46.1\% in SAM-B) cluster in the positive peak and the remaining channels cluster in the negative peak. 

Recently, some methodologies \cite{wei2023outlier,li2023repq} have been introduced to overcome this channel-wise asymmetry by equivalently adjusting the weights of LayerNorm and the subsequent linear. However, we observe that the bimodal distribution does not exist in \cls{post-LayerNorm} activations but rather prominently concentrates in \cls{post-Key-Linear} activations (Figure \ref{fig:frame}), rendering the aforementioned methods inapplicable. To precisely estimate the influence of bimodal distribution, the following matrix multiplication between $\boldsymbol{Q}$ and $\boldsymbol{K}$ can be formulated as
% Existing methods may allocate precious bits to represent the sparse intervals between the two peaks, leading to severe performance drop on extremely low-bit quantization. In recent methodologies \cite{wei2023outlier,li2023repq}, asymmetric post-LayerNorm activations are eliminated by equivalently adjusting the weights of LayerNorm and the subsequent linear. However, we observe that the bimodal distribution does not exist in \cls{post-LayerNorm} activations but rather prominently manifests in \cls{post-Key-Linear} activations, rendering the aforementioned methods inapplicable. To precisely estimate the influence of bimodal distribution, the following matrix multiplication between $Q$ and $K$ can be formulated as: %matrix
% \jy{Here, $m$ is the number of tokens and $n$ is the feature dimension. Xq, Wq, bq is xxx. Xk Wk, Bk is xxx.}
\begin{equation}
\boldsymbol{Q}\boldsymbol{K}^{\top} = 
\underbrace{(\boldsymbol{X}_q\boldsymbol{W}_q+\boldsymbol{b}_q)}_{\text{normal distribution}}
\underbrace{(\boldsymbol{X}_k\boldsymbol{W}_k+\boldsymbol{b}_k)^{\top}}_{\text{bimodal distribution}},
\end{equation}
where $\boldsymbol{W}\in \mathbb{R}^{m \times n}$ and $\boldsymbol{b}\in \mathbb{R}^{n}$ are the weight and bias of linear (\ie, fully connected layer). $m$ and $n$ are input and output feature dimensions. The subscripts $q$ and $k$ denote the query and key linears, respectively. Note that the matrix multiplication of $\boldsymbol{Q}$ and $\boldsymbol{K}$ essentially represents the matrix multiplication of normal and bimodal distributions. Motivated by the above analysis, we adopt a channel-wise sign factor $\boldsymbol{\gamma} \in \mathbb{R}^{n}$ to transfer the bimodal distribution to a normal distribution. We assume that $\boldsymbol{\gamma}$ can be computed from the mean value of each channel, considering the sign factor of $j$-th channel:
\begin{equation}
\label{eq:gamma}
    \boldsymbol{\gamma}_{j}=
    \begin{cases} 
    +1, & \text{if} \,\,\, \mathtt{mean}(\boldsymbol{K}_{:,j}) \geq 0 \\
    -1, & \text{otherwise}
    \end{cases},
\end{equation}
where $\boldsymbol{K}_{:,j}$ denotes the \cls{post-Key-Linear} activations in $j$-th channel. And we compute the $\boldsymbol{\gamma}_{j}$ through its sign of mean value. Specifically, if the $j$-th channel is negative peak, $\boldsymbol{\gamma}_{j}$ will be -1. Then we multiply it to the corresponding channel of query linear and key linear simultaneously to maintain equivalence. After that, $\boldsymbol{K}_{:,j}$ will be transferred to the positive. Conversely, if one channel is positive peak, it will remain invariant as its sign factor is 1. As shown in Figure \ref{fig:frame}, after using our BIG strategy, the negative peak in $\boldsymbol{K}$ has been merged into the positive peak, thereby transferring the bimodal distribution to a normal distribution. On the other hand, $\boldsymbol{Q}$ still preserve normal distribution:
\begin{align}
\label{eq:convert}
\boldsymbol{Q}\boldsymbol{K}^{\top} &=   ((\boldsymbol{X}_q\boldsymbol{W}_q+\boldsymbol{b}_q)\odot \boldsymbol{\gamma})
((\boldsymbol{X}_k\boldsymbol{W}_k+\boldsymbol{b}_k)^{\top} \odot \boldsymbol{\gamma}^{\top}) \notag  \\
&= \underbrace{(\boldsymbol{X}_q\boldsymbol{W}_q^{'}+\boldsymbol{b}_q^{'})}_{\text{normal distribution}}
\underbrace{(\boldsymbol{X}_k\boldsymbol{W}_k^{'}+\boldsymbol{b}_k^{'})^{\top}}_{\text{normal distribution}}.
\end{align}
% \jy{If the distribution in one channel has a negative peak, $\gamma$ will be xxx, and the distribution of this channel will be transfer to the positive. On the other hand, if the distribution in one channel has a positive peak, $\gamma$ will xxx. In this way, we can safely transfer the channels with negative distribition to the positive one, and eliminate the bimodal distribution in this tensor.}

Note the sign factor $\boldsymbol{\gamma}$ can be easily absorbed into previous query linear and key linear offline without any computation overhead, \ie, $\boldsymbol{W}^{'} =\boldsymbol{W} \odot \boldsymbol{\gamma}$ and $\boldsymbol{b}^{'} =\boldsymbol{b} \odot \boldsymbol{\gamma}$. 

\noindent \textbf{bimodal discovery:} However, not all \cls{post-Key-Linear} activations in SAM is bimodal distribution. To discriminate the bimodal distribution, we first adopt the Gaussian kernel density estimation to compute the probability density function (PDF) \cite{scott1992multivariate} by the whole tensor. Based on the continuous and smooth function, we quantitatively describe the peaks as local maxima. To avoid recognizing two small bumps as two peaks, we constrain the peak height and the distance between two peaks. More implementation details are described in supplementary materials.

In summary, our \textit{Bimodal Integration} (BIG) strategy comprises three steps: bimodal discovery, $\boldsymbol{\gamma}$ computation and equivalent transformation. Due to the strong asymmetry, only one sample is enough to compute the sign factor $\boldsymbol{\gamma}$, described in Algorithm \ref{algo:pipeline}. Therefore, our BIG is efficient and the extra computational burden can be ignored.

% To address it, we first adopt the Gaussian kernel density estimation to compute the probability density function (PDF) of the distribution in a non-parametric way. Then, we quantitatively describe the peaks as local maxima and employ corresponding variables spacing to signify the void/sparse intervals amidst the peaks. If exactly two suitable peaks can be found in the entire distribution corresponding to reasonable interval, we address this is a bimodal distribution that we need. At the same time, we verified the previously mentioned properties.
% Then, we quantitatively describe the peaks as local maxima and employ corresponding variable spacing to signify the vacant or sparsely populated intervals amidst the peaks.

% \begin{figure}[t]
% \begin{center}
%      \includegraphics[width=1\linewidth]{figure/sign1.pdf}
% \end{center}
%  \captionsetup{skip=0pt}
%   \caption{sign operation.}
 
%     \label{fig:sign}
% \end{figure}

% \input{sec/sub_sec/discriminator}

\begin{algorithm}\small
\caption{\textsc{PTQ4SAM CALIBRATION}}    
% \label{alternative-pipeline} 
\label{algo:pipeline}
\begin{algorithmic}[1]
\REQUIRE Full-precision SAM $M_F$, Calibration Set $Calib$
% \\ Number of Samples $K$
\ENSURE Quantized model $M_Q$

% \STATE \bluecomment{Calibration}
\FOR{$i = 1: \#Calib$}
\STATE \bluecomment{Bimodal Integration}
\IF{$i == 1$}
\STATE Discriminate bimodal distribution
\STATE Compute sign factor $\boldsymbol{\gamma}$ with Eq.~\ref{eq:gamma}
\STATE Equivalent transformation with Eq.~\ref{eq:convert}
\ENDIF

\FOR{$l=1:\#Quantizer$}

\STATE Initialize $l$-th quantizer's scaling factor 
\STATE \bluecomment{For post-Softmax Distribution}
\IF{Adaptive Granularity Quantizer}
\STATE Compute and add error of each $\tau$
% \ELSE
% \STATE Initialize scaling factor with with Eq.~\ref{eq:search-a}
\IF{$i==\#Calib$}%$i==Len(D_c)-1$
\STATE Determine optimal $\tau$ with Eq.~\ref{eq:search-a}
\ENDIF
\ENDIF
\ENDFOR
% \STATE Update quantizer of $Q$ \textcolor{blue}{\Comment{calibration}}
% \STATE Compute Frobenius norm for $\alpha$ with Eq.~\ref{eq:search-a}
\ENDFOR
% \STATE \bluecomment{Block Reconstruction}
% \FOR{$b=1:\#Block$}
% % \STATE Optimize  $b$-th block through block reconstruction
% \STATE Reconstruct the $b$-th block 
% \IF{\textsc{PTQ4SAM-L}}
% \STATE Block Reconstruction
% \ENDIF
% \ENDFOR
\RETURN $M_Q$
\end{algorithmic}
\end{algorithm}

\subsection{Adaptive Granularity Quantization}
\label{sec:log}

% \begin{figure}[t]
% \begin{center}
%      % \includegraphics[width=1\linewidth]{figure/post-softmax.pdf}
%       \includegraphics[width=1\linewidth]{figure/method_log.pdf}
% \end{center}
%  \captionsetup{skip=0pt}
%   \caption{The histograms of post-Softmax activations of different layers in SAM-L. The x-axis is logarithmic scale.}
%     \label{fig:softmax_show}
% \end{figure}

In transformer architecture, the softmax function is utilized to convert attention scores into probabilities, ensuring that the scores are normalized and lie between 0 and 1. \cite{yuan2022ptq4vit,ding2022towards,li2023repq,lin2021fq} identify the extremely unbalanced power-law distribution as the rationale of quantization difficulty and devise a specialized quantizer to address it. However, the aforementioned methods are mainly designed for the softmax in self-attention mechanism. SAM also incorporates cross-attention in two directions, \ie, \cls{token-to-image} and \cls{image-to-token} cross-attention, amplifying the remarkable discrepancy between post-softmax distributions in Figure \ref{fig:intro_b}. For example, there are more ultra-low-values in \cls{token-to-image}, displaying a smooth distribution under a logarithmic scale. On the contrary, the distributions in \cls{image-to-token} and \cls{self-attention} exhibit higher kurtosis and exist more high values.

% intuitively, there are evident differences in both mean and variance between the these distributions.

To tackle this discrepancy, we revisit the logarithmic quantizer and propose an \textit{Adaptive Granularity Quantization} (AGQ) with an adaptive parameter $\tau$ to adjust the base. As shown in Figure \ref{fig:frame}, a smaller $\tau$ can represent lower attention scores. And as $\tau$ increases, the higher attention scores become more fine-grained. Our AGQ, equipped with a suitable $\tau$, achieves a flexible trade-off between the granularity of low and high values under diverse post-Softmax scenarios and different bit-widths. The corresponding quantization and de-quantization operations of AGQ for attention score $a$ (one element in the attention map for simplicity) can be rewritten as:
\begin{gather}
    a_{q} = \mathtt{clamp}(\lfloor - \log_{ 2^{ \frac{1}{\tau} }} \frac{a}{s} \rceil,0,2^k-1),\\
    \hat{a} = s_a \cdot 2^{- \frac{{a}_{q}}{\tau}}.
\end{gather}

In our implementation, we use $\tau \in \{ 2^0, 2^1,\ldots,2^n\}$ for the hardware efficiency, which will be demonstrated later. The premise for executing the bit-shifting operation is the assurance that the exponential term is an integer. Regrettably, $\frac{{a}_{q}}{\tau}$ is not necessarily an integer. Consequently, we first decompose into its integral and fractional components:% , given by
% As we focus on hardware-efficient quantization for practical usage, we need to assure the exponential term is an integer for  bit-shifting operation.

\begin{equation}
    \hat{a} = s_a\cdot 2^{\lfloor - \frac{{a}_{q}}{\tau} \rfloor} \cdot 2 ^ {\frac{(-{a}_{q}) \% \tau}{\tau}},
\end{equation}
where $\%$ denotes the modulo function and $\lfloor \cdot \rfloor$ denotes the floor function. Subsequently, the multiplication between quantized attention score $\hat{a}$ and value $\hat{v}$ are written as:

\begin{align}
    \hat{a} \cdot \hat{v} &= s_a \cdot s_v \cdot 2 ^ {\frac{(-{a}_{q}) \% \tau}{\tau}}\cdot 2^{\lfloor - \frac{{a}_{q}}{\tau} \rfloor} \cdot v_q \\ 
    &= s_a \cdot s_v \cdot \underbrace{2 ^ {\frac{(-{a}_{q}) \% \tau}{\tau}}}_{(12-1)} \cdot  v_q >>  \lceil \frac{{a}_{q}}{\tau} \rceil,
\end{align}
where $>>$ indicates the bit-shifting operations. As analyzed above, the term (12-1) is an uncertain floating-point value, it limits the efficient integer-only arithmetic in hardware. However, the term (12-1) possesses only $\tau$ discrete values so a small lookup table can be used to avoid it: % , described as
\begin{equation}
    \label{eq:lut}
    \hat{a} \cdot \hat{v} = s_a \cdot s_v  \cdot LUT((-{a}_{q}) \% \tau,v_q >>  \lceil \frac{{a}_{q}}{\tau} \rceil),
\end{equation}
where $LUT$ denotes the small lookup table, which is available on various Neural Network Accelerators, \eg on FPGA. The entries in the LUT are only determined by $(-{a}_{q}) \% \tau$ and $v_q >>  \lceil \frac{{a}_{q}}{\tau} \rceil$. The first term can be represented with an $n$-bit number, and the second term with a $k$-bit (activation bit-width). Notably, since the LUTs for $\tau \in \{ 2^0, 2^1,\ldots,2^{n-1}\}$ can be incorporated into the LUT for $\tau=2^{n}$, the entire network only requires a single LUT. Considering a scenario with 8-bit activation and maximum $\tau$ from the set ($2^2$ for implementation), the size of the LUT is computed as $2^{8+2} \times$4 bytes = 4KB, which is negligible compared to quantized SAMs.
% Thus the number of LUT entries is up to $2^{n+k}$. 

% The first term can be represented using a $n$-bit number and the second term can be represented by a $k$-bit (activation bit-width) number.
%\cite{vogel2018efficient} found that a small size is sufficient for $LUT$. Since $\tau$ is a power of two, the only additional hardware overhead, modulo computation, is negligible by checking the last $n$ bits. Following this, the output of matrix multiplication can be efficiently calculated by accumulation. 

%  in Table~\ref{table:toy_mse}
With the theoretical validation established, our aim is to define an objective to select the optimal $\tau$. To this end, a natural choice is to minimize the local quantization error of attention map $\boldsymbol{A}$ directly. However, we discover it is inconsistent with the quantization error of its associated attention block, which induces instabilities in the global quantization performance, especially at low-bit (see more details in supplementary materials). Therefore, to alleviate this inconsistency, we design the objective function to measure the quantization error of the matrix multiplication output between attention map $\boldsymbol{A}$ and values $\boldsymbol{V}$:
% Rather than directly computing the quantization error of attention map $\boldsymbol{A}$, the objective function measures the quantization error of the matrix multiplication output between attention map $\boldsymbol{A}$ and values $\boldsymbol{V}$, expressed as:
% Formally, the objective function is defined as:
\begin{equation}
\label{eq:search-a}
    \arg\min\limits_{\tau} \mathbb{E}\big[\| \boldsymbol{A} \boldsymbol{V} - \boldsymbol{\widehat{A}}_{\tau} \boldsymbol{V} \|_F^2\big],
\end{equation}
where $\boldsymbol{\widehat{A}}_{\tau}$ represents the quantized attention map by our AGQ with the base $2^{\frac{1}{\tau}}$ and $\| \cdot \|_F^2$ denotes the Frobenius norm to measure the difference. As elaborated in Algorithm \ref{algo:pipeline}, we sort total quantization errors of each $\tau$ across the calibration set, then choose the optimal $\tau$ after calibration.

\begin{table*}
\tiny
\centering
\setlength{\tabcolsep}{2mm}
\resizebox{ \linewidth}{!}{
\begin{tabular}{ll|c|c|c|c|c|c|c|c|c}
% \toprule
% \cmidrule(r){3-5} \cmidrule(r){6-8} \cmidrule(r){9-11}
% \cline{3-11} \rule{0pt}{6pt}
\toprule
% \rule{0pt}{6pt}
 \multirow{2}{*}[-0.8mm]{Detector}    & 
\multirow{2}{*}[-0.8mm]{Methods}  & \multicolumn{3}{c|}{\textbf{SAM-B}}    & \multicolumn{3}{c|}{\textbf{SAM-L}}        & \multicolumn{3}{c}{\textbf{SAM-H}}        \\
% \cmidrule(l{2pt}r{2pt}){3-5} \cmidrule(l{2pt}r{2pt}){6-8} \cmidrule(l{2pt}r{2pt}){9-11} 
\cmidrule(){3-11} 
                             &                          & FP                    & W6A6 & W4A4 & FP                    & W6A6     & W4A4 & FP                    & W6A6     & W4A4 \\ \midrule
\multirow{8}{*}{Faster R-CNN \cite{ren2015faster}} &   MinMax~\cite{jacob2018quantization}             & \multirow{8}{*}{33.4}   &   9.2    &   -    & \multirow{8}{*}{36.4} & 32.9 &   -     & \multirow{8}{*}{37.2}   & 31.9  &   -    \\ &  Percentile~\cite{wu2020integer}                   &                       &    10.9   &  -     &                       &    33.5       &  -     &                       &     32.0      &  -     \\
&  OMSE~\cite{choukroun2019low}                  &                       &     11.9   &  -     &                       &    33.9       &  5.4     &                       &     33.1      &  7.4     \\
&  \cellcolor[HTML]{F3F3F3}\textbf{PTQ4SAM-S}                  &                       &    \cellcolor[HTML]{F3F3F3}\textbf{15.4}   & \cellcolor[HTML]{F3F3F3}-     &                       &    \cellcolor[HTML]{F3F3F3}\textbf{35.7}       &  \cellcolor[HTML]{F3F3F3}\textbf{18.1}     &                       &     \cellcolor[HTML]{F3F3F3}\textbf{36.0}      &  \cellcolor[HTML]{F3F3F3}\textbf{24.1}     \\  
&          AdaRound~\cite{nagel2020up}           &                       &    23.1   &  -     &                       &    34.3       &  8.7     &                       &     33.7      &  14.5     \\
                             & BRECQ~\cite{li2021brecq}                    &                      &   24.1    &   -    &                       &  34.2        &    10.7   &                       &   33.7         &  15.1     \\
                             & QDrop~\cite{wei2022qdrop}                    &                       &   29.3    &   13.0    &                       &    35.2       &  22.6     &                       &      36.3     &   32.3    \\
                             & \cellcolor[HTML]{F3F3F3}\textbf{PTQ4SAM-L}                    &                       & \cellcolor[HTML]{F3F3F3}\textbf{30.3}  & \cellcolor[HTML]{F3F3F3}\textbf{16.0}  &                       &    \cellcolor[HTML]{F3F3F3}\textbf{35.8}       & \cellcolor[HTML]{F3F3F3}\textbf{28.7}    &                       &    \cellcolor[HTML]{F3F3F3}\textbf{36.5}       &   \cellcolor[HTML]{F3F3F3}\textbf{33.5}    \\ \midrule
\multirow{8}{*}{YOLOX \cite{ge2021yolox}}       &   MinMax~\cite{jacob2018quantization}             & \multirow{8}{*}{37.0}   &   10.7    &   -    & \multirow{8}{*}{40.4} & 37.5 &   -     & \multirow{8}{*}{41.0}   & 36.1  &   -    \\ &  Percentile~\cite{wu2020integer}                   &                       &    12.0   &  -     &                       &    38.0       &  -     &                       &     36.3      &  -     \\
&  OMSE~\cite{choukroun2019low}                 &                       &    13.5   &  -     &                       &    38.4       &  6.1     &                       &     37.5      &  7.8     \\
&   \cellcolor[HTML]{F3F3F3}\textbf{PTQ4SAM-S}                  &                       &    \cellcolor[HTML]{F3F3F3}\textbf{17.4}   &  \cellcolor[HTML]{F3F3F3}-     &                       &    \cellcolor[HTML]{F3F3F3}\textbf{40.0}       &  \cellcolor[HTML]{F3F3F3}\textbf{20.6}     &                       &     \cellcolor[HTML]{F3F3F3}\textbf{40.3}      &  \cellcolor[HTML]{F3F3F3}\textbf{26.7}     \\
&          AdaRound~\cite{nagel2020up}           &                       &    26.4   &  -     &                       &    38.9       &  11.1     &                       &     38.3      &  16.7     \\
                             & BRECQ~\cite{li2021brecq}                    &                       &    26.1   &  -     &                       &    38.9       &  12.0     &                       &     38.3      &  16.3     \\
                             & QDrop~\cite{wei2022qdrop}                    &                       &      33.6  &   13.3    &                       &   39.7        &  25.3     &                       &       40.4    &  35.8     \\
                             & \cellcolor[HTML]{F3F3F3}\textbf{PTQ4SAM-L}                   &                       & \cellcolor[HTML]{F3F3F3}\textbf{34.3}  & \cellcolor[HTML]{F3F3F3}\textbf{18.4}  &                       &    \cellcolor[HTML]{F3F3F3}\textbf{40.3}       & \cellcolor[HTML]{F3F3F3}\textbf{31.6} &                       &   \cellcolor[HTML]{F3F3F3}\textbf{40.7}        &   \cellcolor[HTML]{F3F3F3}\textbf{37.6}    \\ \midrule
\multirow{8}{*}{H-Deformable-DETR \cite{jia2023detrs}}        &   MinMax~\cite{jacob2018quantization}             & \multirow{8}{*}{38.2}   &   10.9    &   -    & \multirow{8}{*}{41.5} & 38.6 &   -     & \multirow{8}{*}{42.0}   & 37.3  &   -    \\ &  Percentile~\cite{wu2020integer}                   &                       &    12.3   &  -     &                       &    39.0       &  -     &                       &     37.5      &  -     \\
&  OMSE~\cite{choukroun2019low}                  &                       &    15.0   &  -     &                       &    39.6       &  6.2     &                       &     38.6      &  7.7    \\
&   \cellcolor[HTML]{F3F3F3}\textbf{PTQ4SAM-S}                  &                       &    \cellcolor[HTML]{F3F3F3}\textbf{17.9}   &  \cellcolor[HTML]{F3F3F3}-     &                       &    \cellcolor[HTML]{F3F3F3}\textbf{41.0}       &  \cellcolor[HTML]{F3F3F3}\textbf{20.9}     &                       &     \cellcolor[HTML]{F3F3F3}\textbf{41.3}      &  \cellcolor[HTML]{F3F3F3}\textbf{27.3}     \\
&          AdaRound~\cite{nagel2020up}           &                       &    27.2   &  -     & &39.9 &8.0 & &                    39.4      &  16.3     \\
                             & BRECQ~\cite{li2021brecq}                    &                       &   27.9    &  -     &                       &     39.9      &   11.1    &                       &    39.5       &   15.5    \\
                             & QDrop~\cite{wei2022qdrop}                    &                       &   34.3    &    13.2   &                       &     40.5      &  25.8     &                       &     41.4      &   36.5    \\
                             & \cellcolor[HTML]{F3F3F3}\textbf{PTQ4SAM-L}                      &                       &    \cellcolor[HTML]{F3F3F3}\textbf{35.1}   &   \cellcolor[HTML]{F3F3F3}\textbf{17.3}    &                       &  \cellcolor[HTML]{F3F3F3}\textbf{41.2}         &    \cellcolor[HTML]{F3F3F3}\textbf{32.1}   &                       &    \cellcolor[HTML]{F3F3F3}\textbf{41.6}       &   \cellcolor[HTML]{F3F3F3}\textbf{38.4}    \\ \midrule
\multirow{8}{*}{DINO \cite{zhang2022dino}}        &   MinMax~\cite{jacob2018quantization}             & \multirow{8}{*}{44.5}   &   11.2    &   -    & \multirow{8}{*}{48.6} & 44.7 &   -     & \multirow{8}{*}{49.1}   & 42.8  &   -    \\ &  Percentile~\cite{wu2020integer}                   &                       &    14.0   &  -     &                       &    45.4       &  -     &                       &     43.1      &  -     \\
&  OMSE~\cite{choukroun2019low}                  &                       &    16.6   &  -     &                       &    45.9       &  6.8     &                       &     44.5      &  8.3     \\
&   \cellcolor[HTML]{F3F3F3}\textbf{PTQ4SAM-S}                  &                       &    \cellcolor[HTML]{F3F3F3}\textbf{20.4}   &  \cellcolor[HTML]{F3F3F3}-     &                       &   \cellcolor[HTML]{F3F3F3}\textbf{47.7}       &  \cellcolor[HTML]{F3F3F3}\textbf{23.1}    &                       &     \cellcolor[HTML]{F3F3F3}\textbf{48.1}      &  \cellcolor[HTML]{F3F3F3}\textbf{30.5}    \\
&          AdaRound~\cite{nagel2020up}           &                       &    31.2   &  1.2     &                       &    46.6       &  8.8     &                       &     46.0      &  18.2     \\
                             & BRECQ~\cite{li2021brecq}                    &                       &    31.8   & 3.6        &                       & 46.6           &       12.3 &                       & 46.0         &  17.6      \\
                             & QDrop~\cite{wei2022qdrop}                    &                       &   38.9    &   11.2    &                       &   47.5        &   27.5    &                       &   48.3        &  41.7     \\
                             & \cellcolor[HTML]{F3F3F3}\textbf{PTQ4SAM-L}                      &                     &  \cellcolor[HTML]{F3F3F3}\textbf{40.4}       &    \cellcolor[HTML]{F3F3F3}\textbf{14.4}   &                      & \cellcolor[HTML]{F3F3F3}\textbf{48.3}          &   \cellcolor[HTML]{F3F3F3}\textbf{36.6}    &                       &    \cellcolor[HTML]{F3F3F3}\textbf{48.7}       &   \cellcolor[HTML]{F3F3F3}\textbf{43.9}    \\ \bottomrule
\end{tabular}
}

\caption{Quantization results of instance segmentation on COCO dataset among different detectors. We provide PTQ4SAM-S and PTQ4SAM-L for both statistic-based and learning-based version. - indicates the final result (mAP) is below 1.}
\label{table:main}
\vspace{-1mm}
\end{table*}

% * represents our PTQ4SAM plugged into OMSE~\cite{choukroun2019low} and $\dag$ represents our PTQ4SAM plugged into QDrop~\cite{wei2022qdrop}.
\section{Experiments}
\subsection{Experimental Setup}
\textbf{Tasks, datasets and metrics.} We conduct experiments on three mainstream vision tasks. For the instance segmentation task, we utilize predicted boxes generated by the detector as box prompts for SAM to gain accurate binary masks and evaluate its effectiveness on MS-COCO \cite{lin2014microsoft} dataset with the metric mean Average Precision (mAP). For semantic segmentation task, the overall framework comprises two branches, and we leverage the fine-grained mask produced by SAM to refine the blurry and imprecise mask boundaries generated by the original segmentor. We evaluate its effectiveness on ADE20K \cite{zhou2017scene} dataset using the mean Intersection over Union (mIOU) as the performance metric. For oriented object detection task, we obtain the final rotated RBoxes by the minimum circumscribed rectangle operation on the masks generated by SAM. Our evaluation of its effectiveness on the DOTA-v1.0 \cite{xia2018dota} dataset uses mAP.
% For semantic segmentation task, we leverage the fine-grained mask produced by SAM to refine the blurry and imprecise mask boundaries generated by the original segmentor. We evaluate its effectiveness on ADE20K \cite{zhou2017scene} dataset and Cityscapes \cite{cordts2016cityscapes} dataset.

\noindent
\textbf{Implementation details.} We choose CNN-based Faster R-CNN \cite{ren2015faster}, YOLOX \cite{ge2021yolox}, FCOS \cite{tian2019fcos} and transformer-based H-Deformable-DETR \cite{jia2023detrs}, DINO \cite{zhang2022dino} as detectors and advanced SegFormer \cite{xie2021segformer} as segmentor. The adaptive parameter $\tau$ is searched from the set $\{ 2^0, 2^1, 2^2\}$. We set the box threshold to 0.05 for CNN-based detectors and affix a set of 100 adaptive anchors for transformer-based detectors. We randomly sample 32 training images as calibration set and only the first sample is utilized for the determination of the bimodal distribution. For a fair comparison, we adopt per-channel asymmetric quantization for weights and per-tensor asymmetric quantization for activations~\cite{wei2022qdrop,liu2023pd}. Following the common settings~\cite{yuan2022ptq4vit,wei2022qdrop}, the first and last layer/block are not quantized. To verify the effectiveness of PTQ4SAM in two kinds of PTQ methods, we integrate our method into statistic-based OMSE \cite{choukroun2019low} and learning-based QDrop \cite{wei2022qdrop}, called PTQ4SAM-S and PTQ4SAM-L, respectively. For learning-based methods, we adopt MinMax calibration strategy and design attention block,  MLP block for block-wise reconstruction with 20000 iterations.
% Our framework is implemented in PyTorch, and all experiments are conducted on 4 NVIDIA Tesla A100 GPUs with 80 GB  memory.

\subsection{Instance Segmentation Results}
We compare our PTQ4SAM with statistic-based methods such as MinMax \cite{jacob2018quantization}, Percentile \cite{wu2020integer}, OMSE \cite{choukroun2019low} and learning-based methods such as AdaRound \cite{nagel2020up}, BRECQ \cite{li2021brecq}, QDrop \cite{wei2022qdrop}. Table \ref{table:main} lists the performance of all methods. Our method consistently outperforms other methods by a large margin among different detectors. Our 4-bit PTQ4SAM-S achieves comparable results, and exceeds the baseline OMSE by over 10\% mAP (\eg, from 5.4\% to 18.1\% with Faster R-CNN) on SAM-L and about 20\% mAP (\eg, from 8.3\% to 30.5\% with DINO) on SAM-H, recovering them to a usable level. Our PTQ4SAM-S even remarkably surpasses the state-of-the-art learning-based method QDrop, at W6A6 on SAM-L. Meanwhile, our PTQ4SAM-L encouragingly achieves lossless accuracy. For instance, at W6A6 setting, our PTQ4SAM-L achieves 40.3\% and 41.2\% when applying YOLOX and H-Deformable-DETR on SAM-L, with only 0.1\% and 0.3\% performance drop compared to full-precision models. When quantizing to more challenging case W4A4, AdaRound and BRECQ become infeasible while our method surpasses the QDrop by 5.1\% on SAM-B with YOLOX and 6.3\% on SAM-L with H-Deformable-DETR. Compared with SAM-B and SAM-L, SAM-H exhibits greater robustness when introducing quantization noise, but our PTQ4SAM-L still provides about 2\% improvement at W4A4. Applying the state-of-art detector DINO \cite{zhang2022dino}, our 6-bit PTQ4SAM-L yields 40.4\% mAP on SAM-B and achieves a mAP nearing 50\% on SAM-L and SAM-H.
% (\eg, 20.6\% and 26.7\% on SAM-L and SAM-H with YOLOX)
% We compare our PTQ4SAM with the state-of-the-art methods such as AdaRound \cite{nagel2020up}, BRECQ \cite{li2021brecq} and QDrop \cite{wei2022qdrop}. Table \ref{table:main} lists the performance of all methods. Our method outperforms other methods by a large margin among different detectors. 
% Furthermore, to validate its effectiveness against the current mainstream instance segmentor, we choose the powerful DINO \cite{zhang2022dino} as the detector of SAM.
\subsection{Semantic Segmentation Results}

\begin{table}[t]
\small
\centering
\setlength{\tabcolsep}{2mm}
\resizebox{\linewidth}{!}{

\begin{tabular}{c|c|l|c|cc}
\toprule
\textbf{Base} & \textbf{Model}           & \textbf{Methods}              & \textbf{FP}           & \textbf{W6A6} & \textbf{W4A4} \\ \midrule
\multirow{12}{*}[-0.1mm]{\textbf{31.78}} & \multirow{4}{*}{\textbf{+SAM-B}} & AdaRound~\cite{nagel2020up} & \multirow{4}{*}{33.15} & 32.34               & 31.78               \\
                    &   & BRECQ~\cite{li2021brecq}    &                       & 32.27               & 31.78               \\
                   &    & QDrop~\cite{wei2022qdrop}    &                       & 32.57           &   31.79         \\
                 &      & \textbf{PTQ4SAM-L}     &                       & \textbf{32.65}          &  \textbf{31.85}            \\ \cline{2-6}
& \multirow{4}{*}{\textbf{+SAM-L}} & AdaRound~\cite{nagel2020up} & \multirow{4}{*}{33.61} & 32.99               & 31.97             \\
                 &      & BRECQ~\cite{li2021brecq}   &                       & 33.04               & 31.98              \\
                &       & QDrop~\cite{wei2022qdrop}    &                       & 33.58           & 32.67          \\
                  &     & \textbf{PTQ4SAM-L}     &                       & \textbf{33.66}          & \textbf{32.82}         \\  \cline{2-6}
& \multirow{4}{*}{\textbf{+SAM-H}} & AdaRound~\cite{nagel2020up} & \multirow{4}{*}{33.63} & 33.49               & 32.17             \\
             &          & BRECQ~\cite{li2021brecq}    &                       & 33.46               & 32.12             \\
                &       & QDrop~\cite{wei2022qdrop}    &                       & 33.49           & 32.93           \\
               &        & \textbf{PTQ4SAM-L}     &                       & \textbf{33.66}          & \textbf{33.10}   \\ \bottomrule
\end{tabular}

}
\caption{Quantization results of semantic segmentation. SAM refines the outcomes produced by the original semantic segmentor.}
% \vspace{-1mm}
% The framework comprises semantic segmentor and SAM. The segmentor obtains 31.78 mIOU on ADE20K. SAM blurry and imprecise 
\label{table:semantic}
\end{table}

In this part, we verify the performance of the learning-based PTQ methods for semantic segmentation task. The original semantic segmentor achieves 31.78\% on  ADE20K dataset, and SAMs adjust the preliminary outcomes of segmentor, \ie, bring 1.37\%, 1.83\%, 1.85\% enhancement for full-precision SAM-B, SAM-L and SAM-H. As shown in Table \ref{table:semantic}, quantized SAMs generally still contribute to the final masks. Our method retains the capability of SAM to the greatest extent. In particular, we are surprised to find that 6-bit PTQ4SAM-L even achieves better performance than the full-precision models on both SAM-L and SAM-H. At W4A4 setting, our method provides 1.04\% accuracy promotion on SAM-L, outperforming QDrop by 0.15\%.

% consistently yields gratifying performances
% The full-precision SAM-H model can bring about a 1.22\% mAP improvement, while the SAM-L and SAM-H 6-bit quantized by our method can still bring about an accuracy improvement of over 1\%. In INT4 setting, SAM-H quantized with our method can also contribute up to performance improvement of 1.32\%, but qdrop is only 1.15\%. Although SAM itself has limited gain effect on this task, it can also be found that our method always outperforms existing methods.

\subsection{Object Detection Results}

\begin{table}[t]
\small
\centering
\setlength{\tabcolsep}{4mm}
\resizebox{0.48\textwidth}{!}{
\begin{tabular}{l|l|c|cc}
\toprule
\textbf{Model}           & \textbf{Methods}              & \textbf{FP}           & \textbf{W6A6} & \textbf{W4A4} \\ \midrule
\multirow{4}{*}{\textbf{SAM-B}} & AdaRound~\cite{nagel2020up} & \multirow{4}{*}{64.1} & 34.05              & -              \\
                       & BRECQ~\cite{li2021brecq}    &                       & 34.40              & -              \\
                       & QDrop~\cite{wei2022qdrop}    &                       & 59.27          & 41.96          \\
                       &  \textbf{PTQ4SAM-L}     &                       &  \textbf{60.33}          &  \textbf{44.18}             \\ \midrule
\multirow{4}{*}{\textbf{SAM-L}} & AdaRound~\cite{nagel2020up} & \multirow{4}{*}{64.2} & 63.44              & 23.18              \\
                       & BRECQ~\cite{li2021brecq}   &                       & 63.60              & 26.89              \\
                       & QDrop~\cite{wei2022qdrop}    &                       & 63.86          & 50.11         \\
                       &  \textbf{PTQ4SAM-L}     &                       &  \textbf{63.91}          &  \textbf{56.29}         \\ \midrule
\multirow{4}{*}{\textbf{SAM-H}} & AdaRound~\cite{nagel2020up} & \multirow{4}{*}{64.6} & 62.73              & 24.45              \\
                       & BRECQ~\cite{li2021brecq}    &                       & 62.58              & 25.98              \\
                       & QDrop~\cite{wei2022qdrop}    &                       & 62.83          & 55.87          \\
                       & \textbf{PTQ4SAM-L}     &                       &  \textbf{64.36}          &  \textbf{56.01}          \\ \bottomrule
\end{tabular}
}
\caption{Quantization results of oriented object detection.}
\vspace{-1mm}
\label{table:det}
\end{table}

To further demonstrate versatility across other tasks, we test PTQ4SAM-L in oriented object detection. Notably, as shown in Table \ref{table:det}, our method consistently performs better than other learning-based PTQ methods. For instance, when quantizing the network to W6A6, experiments indicate PTQ4SAM-L slightly drops about 0.3\% compared with the full-precision model on SAM-L and SAM-H. At the most challenging W4A4 bit-width, AdaRound and BRECQ suffer from non-trivial performance degradation. Contrastively, our method can still obtain satisfactory performance. We achieve over 44\% and 56\% accuracy on SAM-B and SAM-L, surpassing the baseline QDrop 2.2\% and 6.2\%.
% We also validate the performance of our quantization framework for the SAM model on the oriented object detection task. 
%We adopt FCOS\cite{tian2019fcos} detector with ResNet50\cite{he2016deep} backbone and FPN\cite{lin2017feature} neck to work with the sam model to generate the final rotated RBoxes.%
% The comparison method and bit number settings remain the same as before. The specific results are shown in Table \ref{table:det}. Under different cases, our PTQ framework can always enable the SAM to maintain the best performance. For instance, at INT6 setting, our PTQ4SAM achieves 64.36\% accuracy only 0.2\% performance drop compared to the FP model on SAM-H, whereas the QDrop achieves a 1.8\% drop. In W4/A4 quantization, We achieve over 56\% accuracy in SAM-H, an enhancement of over 6\% compared to other methods, and even quantization for the more refined model SAM-B achieved 44.18\% accuracy. Overall, our approach consistently achieves superior outcomes on oriented object detection tasks. However, in some INT4 settings, AdaRound and BRECQ are almost unfeasible.

\subsection{Ablation Studies}
\textbf{Ablation for components:} Table \ref{tab:ablation_study} lists the results of different components. We demonstrate that each component contributes to PTQ4SAM, while the best performance is achieved when both components are jointly applied. 
Specifically, at relatively higher bit-widths, like W6A6 setting, both BIG and AGQ strategies can bring performance improvement, making the quantized model comparable to the full-precision one. When quantizing SAM to lower bit-widths, \ie, W4A4, BIG can significantly improve the performance, which is 3.4\% higher than baseline, as it can minimize the more serious quantization perturbation.
% Specifically, at a relatively higher bit-widths, like W6A6 setting, AGQ module plays a more important role, which achieves 41.2, indicating that the AGQ can preserve the low values through appropriate trade-off. 

\begin{table}[t]
    \centering
    
    \resizebox{0.44\textwidth}{!}{
    % \Large
    \begin{tabular}{*{1}{c}|*{1}{c}|*{2}{c}|*{1}{c}|*{2}{c}}
        \toprule
        % \multicolumn{1}{c}{} & \multicolumn{1}{c||}{} & \multicolumn{2}{c||}{Synthetic Dataset} & \multicolumn{2}{c}{Static Data}  \\
        \textbf{ Row ID } & \textbf{Model} &\textbf{BIG} & \textbf{AGQ} &\textbf{FP} & \textbf{W6A6} & \textbf{W4A4}  \\
        \midrule
         1& \multirow{4}*{\textbf{SAM-L}} & $\times$ & $\times$ & \multirow{4}*{41.5} & 40.5 & 25.8  \\ %
          2&  & \checkmark & $\times$ &  & 40.6 & 29.2 \\
        % \rowcolor{gray!20}
          3 &  & $\times$ & \checkmark &  & 41.2 & 27.3 \\
        % \midrule
          4&  & $ \checkmark $ & $ \checkmark $ &  & \textbf{41.2}  & \textbf{32.1} \\ %

        \bottomrule
    \end{tabular}
    }
    \caption{Ablation study for key components.} 
    % \vspace{-1mm}
    \label{tab:ablation_study}
    
\end{table}

\begin{table}[t]
    \centering

    \resizebox{0.48\textwidth}{!}{
    \begin{tabular}{c|c|ccccc}
    \toprule
    \textbf{\#bits}  & \textbf{Quantizer} & \textbf{SAM-B} & \textbf{SAM-L} & \textbf{SAM-H} \\ \midrule
    Full-precision  & - & 37.0 &  40.4 & 41.0  \\ \midrule
    
    \multirow{3}*{W6A6}  & Uniform   &  {33.6}   &  39.7  &  40.4     \\ 
    ~ & Log2 & 33.3 & 40.2 & 40.6 \\
    ~ & AGQ (ours) & \textbf{33.9} & \textbf{40.3} & \textbf{40.6} \\    \midrule
    
    \multirow{3}*{W4A4}  & Uniform   &  13.3  & 25.3  &  35.8     \\ 
    ~ & Log2 & 14.1 & 26.5  & 37.3 \\
    ~ & AGQ (ours) & \textbf{15.0} & \textbf{27.8} & \textbf{37.6} \\   
     \bottomrule
    \end{tabular}
    }
    \captionsetup{width=1\linewidth}
    \caption{Ablation study for different post-Softmax quantizers. }
    \label{tab:softmax-quant}
\end{table}

\noindent \textbf{Ablation for different quantizers:} We also report the results of different quantizers in Table \ref{tab:softmax-quant}. Notably, simply employing Log2 quantizer \cite{miyashita2016convolutional} for attention scores is unstable under different settings (\eg, lower result at W6A6 for SAM-B compared with uniform quantizer). Our AGQ, by contrast, surpasses the uniform quantizer and Log2 quantizer in different cases, boosting 3.5\% for 4-bit SAM-L. Apart from the encouraging performance, our AGQ is available on various hardware, ensuring efficient execution.

\begin{figure}[t]
    \centering
    \subcaptionbox{Acceleration \label{fig:speed}}{\includegraphics[width=.365\linewidth]{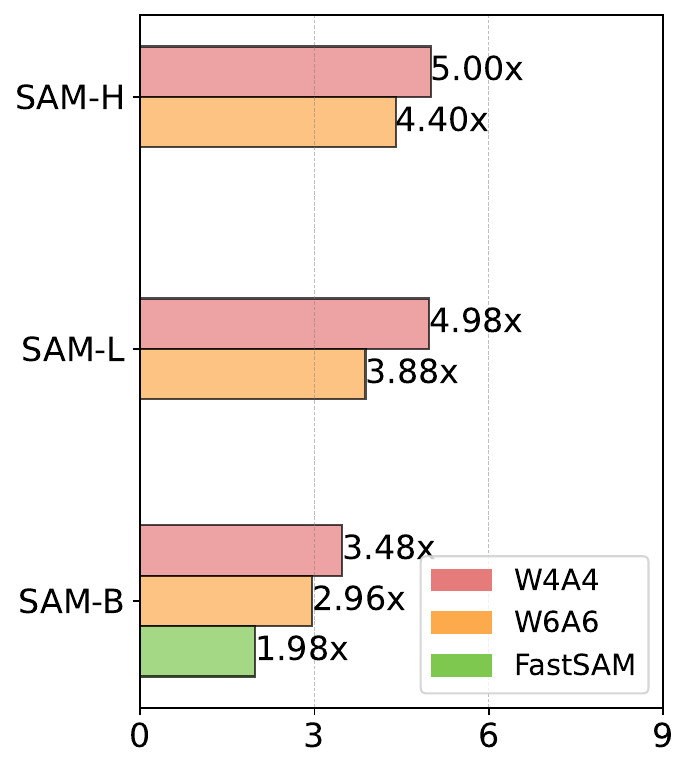}}
    \subcaptionbox{Storage (GB) \label{fig:storage}}{\includegraphics[width=.615\linewidth]{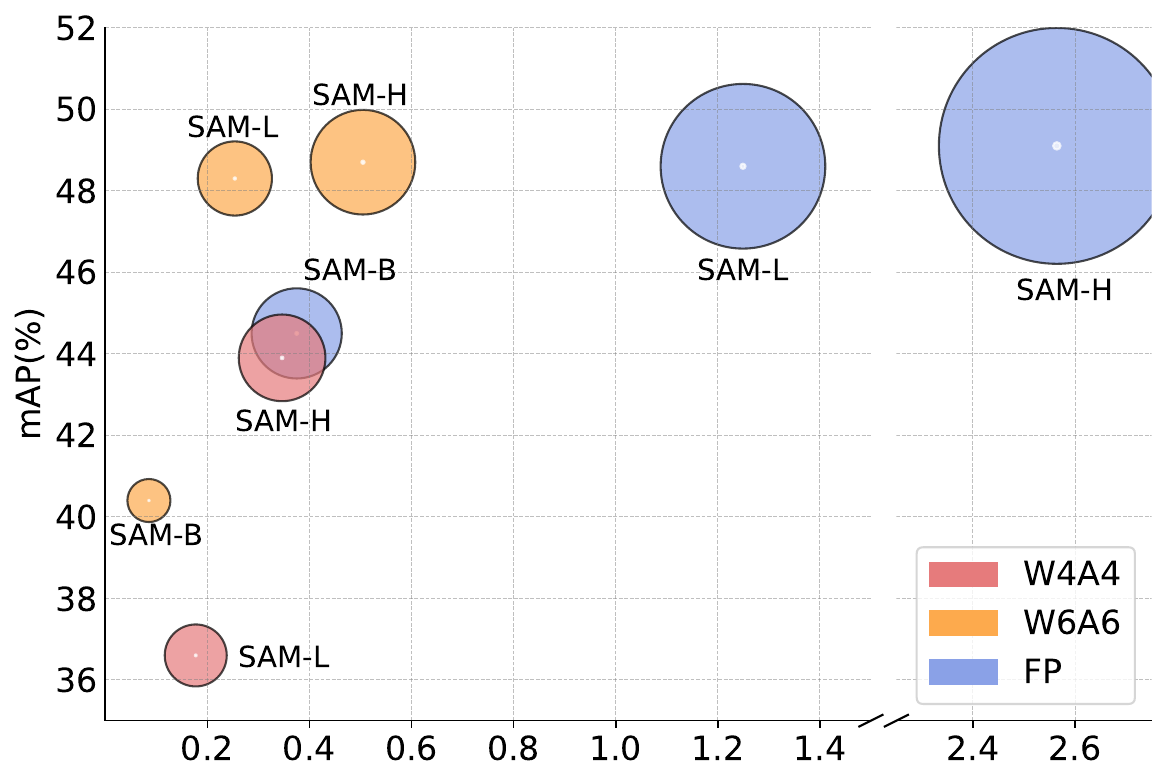}}
    \caption{(a) Theoretical acceleration rate (100 prompts) vs. all SAM models. (b) Accuracy vs. storage. 
    % Circles, rectangles, and triangles mean full-precision, 6-bit and 4-bit models.
    }
    % \label{fig:storage}

\end{figure}

\begin{figure}[t] \centering
    \newcommand{\hwidth}{1pt}
 
    % \makebox[0.15\textwidth]{\small FP} 
    % \\
    \includegraphics[width=0.15\textwidth]{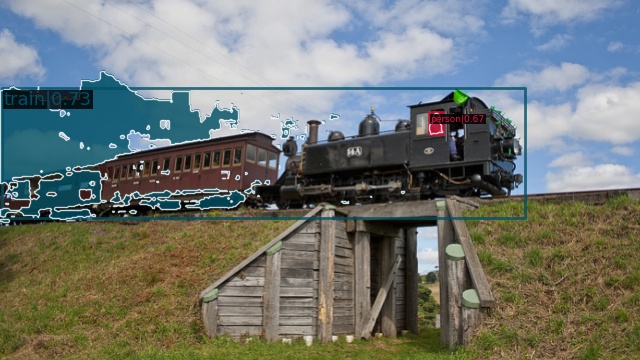}
    \includegraphics[width=0.15\textwidth]{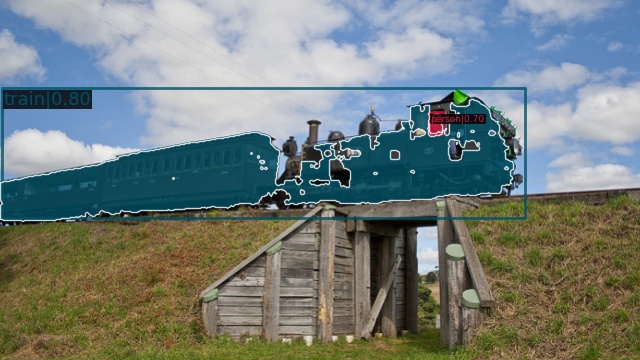}
    \includegraphics[width=0.15\textwidth]{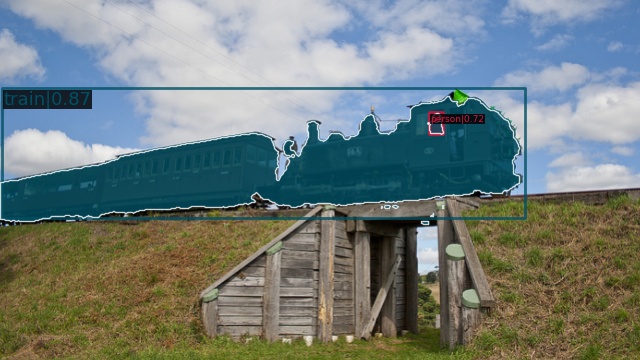}
    \\[2pt]
    % \includegraphics[width=0.15\textwidth]{figure/vis/brecq/000000027982.jpg}
    % \includegraphics[width=0.15\textwidth]{figure/vis/qdrop/000000027982.jpg}
    % \includegraphics[width=0.15\textwidth]{figure/vis/our/000000027982.jpg}
    % % \includegraphics[width=0.15\textwidth]{figure/vis/fp/000000027982.jpg}
    % \\[2pt]
    \includegraphics[width=0.15\textwidth]{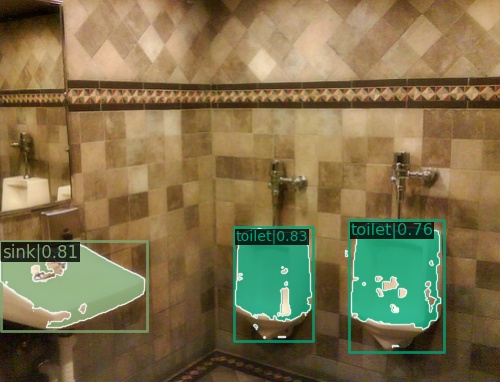}
    \includegraphics[width=0.15\textwidth]{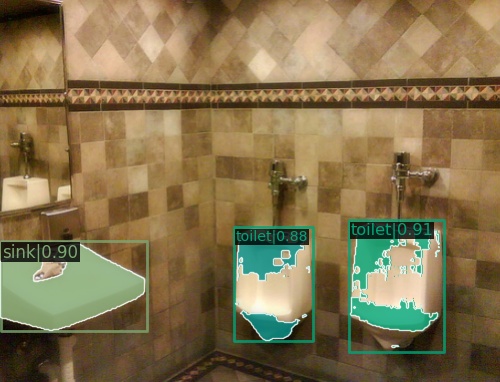}
    \includegraphics[width=0.15\textwidth]{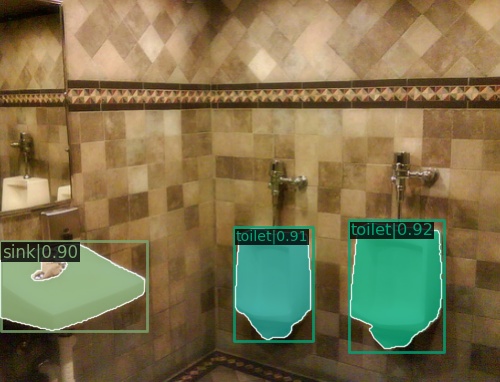}
    \\[2pt]
    \includegraphics[width=0.15\textwidth]{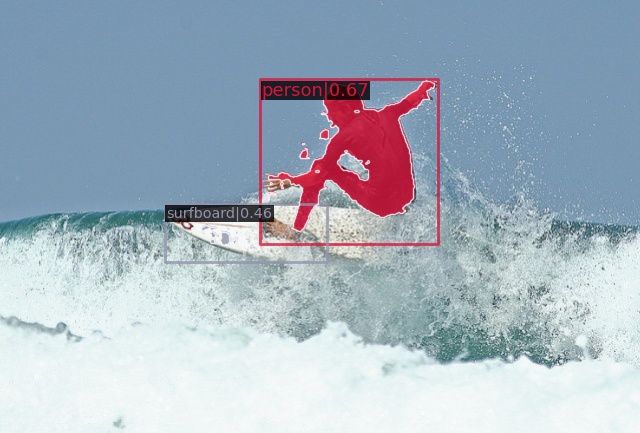}
    \includegraphics[width=0.15\textwidth]{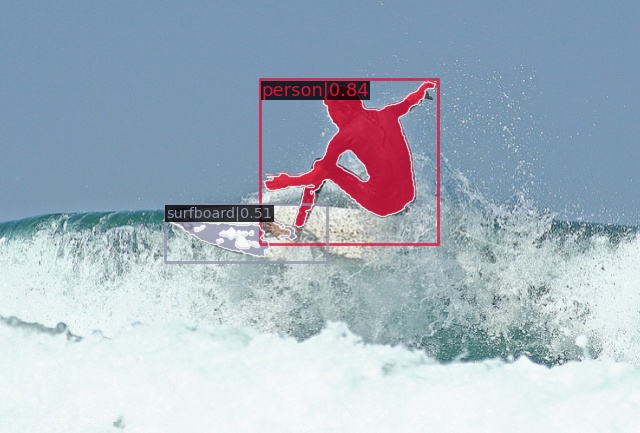}
    \includegraphics[width=0.15\textwidth]{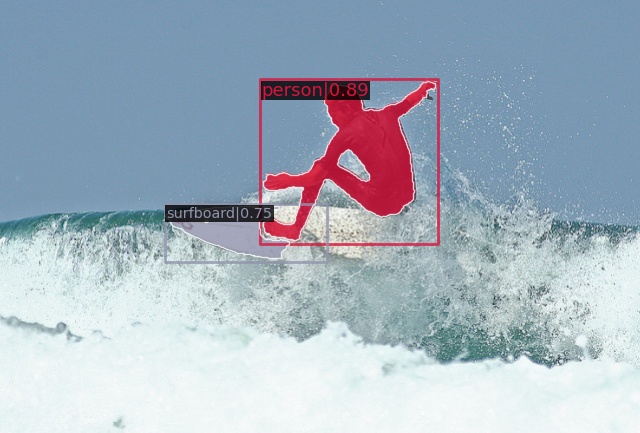}
    \\[2pt]
    \includegraphics[width=0.15\textwidth]{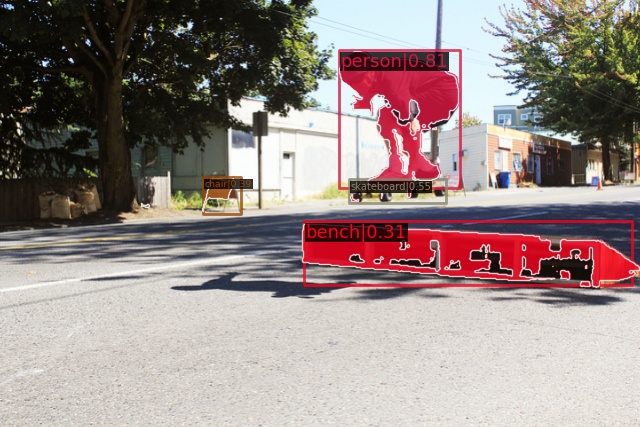}
    \includegraphics[width=0.15\textwidth]{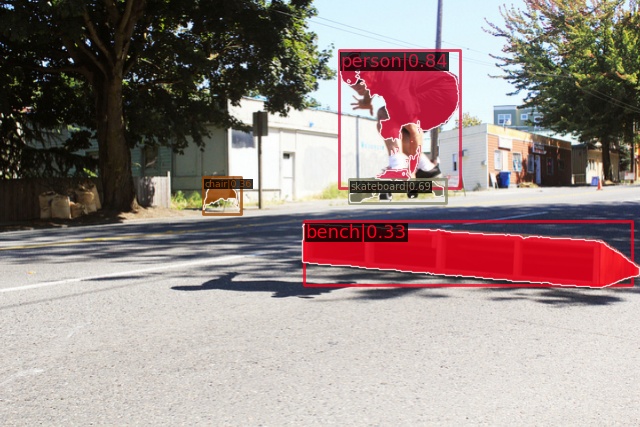}
    \includegraphics[width=0.15\textwidth]{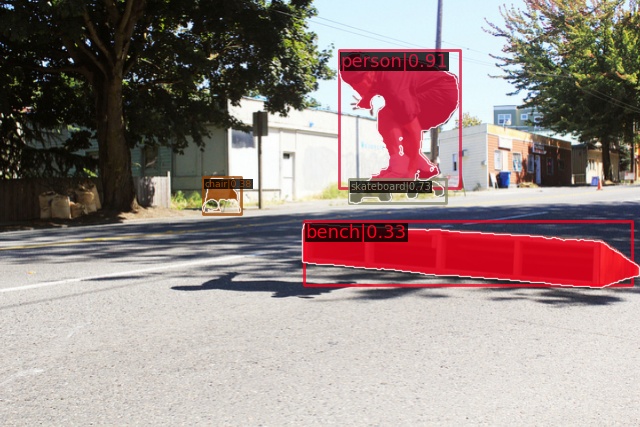}
    \\[2pt]
    % \includegraphics[width=0.15\textwidth]{figure/vis/brecq/000000005060.jpg}
    % \includegraphics[width=0.15\textwidth]{figure/vis/qdrop/000000005060.jpg}
    % \includegraphics[width=0.15\textwidth]{figure/vis/our/000000005060.jpg}
    % \includegraphics[width=0.15\textwidth]{figure/vis/fp/000000005060.jpg}
    % \\[2pt]
    % \includegraphics[width=0.15\textwidth]{figure/vis/brecq/000000007088.jpg}
    % \includegraphics[width=0.15\textwidth]{figure/vis/qdrop/000000007088.jpg}
    % \includegraphics[width=0.15\textwidth]{figure/vis/our/000000007088.jpg}
    % \includegraphics[width=0.15\textwidth]{figure/vis/fp/000000007088.jpg}
    % \\[2pt]
    % \includegraphics[width=0.15\textwidth]{figure/vis/brecq/000000007511.jpg}
    % \includegraphics[width=0.15\textwidth]{figure/vis/qdrop/000000007511.jpg}
    % \includegraphics[width=0.15\textwidth]{figure/vis/our/000000007511.jpg}
    % \includegraphics[width=0.15\textwidth]{figure/vis/fp/000000007511.jpg}
    % \\[2pt]
    \makebox[0.15\textwidth]{\small BRECQ}
    \makebox[0.15\textwidth]{\small QDrop}
    \makebox[0.15\textwidth]{\small Ours}

    % \makebox[0.15\textwidth]{%
    %     \includegraphics[width=0.115\textwidth]{example-image}\hspace{\hwidth}
    %     \includegraphics[width=0.115\textwidth]{example-image}} 
    % \makebox[0.15\textwidth]{%
    %     \includegraphics[width=0.115\textwidth]{example-image}\hspace{\hwidth}
    %     \includegraphics[width=0.115\textwidth]{example-image}}
    % \makebox[0.15\textwidth]{%
    %     \includegraphics[width=0.115\textwidth]{example-image}\hspace{\hwidth}
    %     \includegraphics[width=0.115\textwidth]{example-image}}
    % \makebox[0.15\textwidth]{%
    %     \includegraphics[width=0.115\textwidth]{example-image}\hspace{\hwidth}
    %     \includegraphics[width=0.115\textwidth]{example-image}}
    % \\[-0.1em]
    % \makebox[\textwidth]{\small (a) Results on data A.} 
    % \\[0.5em]

    \caption{Visualization of instance segmentation on 4-bit SAM-L.} 

    \label{fig:fig9}
\end{figure}

\subsection{Storage Saving and Speedup}
We separately calculate the computational complexity in Figure~\ref{fig:speed} and the memory usage in Figure~\ref{fig:storage}. Following~\cite{wang2020bidet,liu2018bi}, we use FLOPs to measure the theoretical acceleration effect, which is equal to the amount of 32-bit floating point multiplication plus \{$\frac{1}{8},\frac{6}{32}$\} of the amount of \{4,6\}-bit multiplications. Surprisingly, our 6-bit SAM-B (2.96$\times$) achieves better acceleration than FastSAM~\cite{zhao2023fast} (1.98$\times$) while maintaining a close performance. At W4A4, our method reduces computational FLOPs by over 70\% and storage by over 85\%. As models scale up, both acceleration ratio and memory savings become more significant, while the performance drop becomes less.
% We separately calculate the storage and computational complexity for the model by counting the number of model parameters and accumulating the computational complexity of each operator. The theoretical acceleration ratio of quantized models can be characterized using the number of floating-point operations (FLOPs). Regarding the quantization aspect, we refer to \cite{wang2020bidet} for processing. In Figure~\ref{fig:storage}, we discover 6-bit PTQ4SAM achieves up to 5$\times$ storage saving and 4$\times$ acceleration. At W4A4, our method reduces computational FLOPs by over 80\% and storage by over 85\%. As models scale up, both acceleration ratio and memory savings become more significant, while the performance drop becomes less.
% To explore memory usage and computation of PTQ4SAM,
% For instance, the size of a 32-bit full-precision floating-point parameter is equivalent to eight 4-bit integers, and the computational complexity of multiplying two 4-bit integers is 1/8 * 1/8 of multiplying two 32-bit floating-point number.
\subsection{Qualitative Results}

% Figure \ref{fig:fig9} shows the visualization results of different quantized SAMs applied to the instance segmentation task. As shown in the figure, our PTQ4SAM quantization framework can better preserve the model's ability to capture object boundaries and ensure the integrity of the object.

To exhibit the superiority of our PTQ4SAM, especially on low-bit quantization (W4A4), we visualize the instance segmentation results in Figure \ref{fig:fig9} compared with other existing SOTA methods on COCO dataset. We can perceive that most methods fail to produce clear boundaries and miss the salient pixels in the center. For example, other methods do not distinguish the complex edge, even erroneously categorizing the sky as foreground (\cls{train} on row 1). Besides, these schemes are unable to detect the objects against complicated backgrounds (\cls{surfboard} on row 3). Furthermore, our method outperforms other methods on ensuring the integrity of the object (\cls{toilet} and \cls{person} on rows 2, 4).
% experiences a non-trivial  performance degradation
% other methods ignore the central region, while our method on.

\section{Conclusion}
In this paper we first propose a novel post-training quantization framework, PTQ4SAM for Segment Anything Model. To begin with, we observe the significant bottleneck lies in bimodal distribution and explore its occurrence position. We introduce Bimodal Integration strategy to eliminate the negative effect of quantization caused by such bimodal distribution. We also present the Adaptive Granularity Quantization which can fit diverse post-Softmax distribution by searching the hardware-friendly base. Extensive experiments demonstrate that our method consistently yields gratifying results across various tasks. Nevertheless, the reason for bimodal distribution in SAM remains unclear. This direction serves as a potential avenue for our future research.

% { \footnotesize
% \textbf{Acknowledgement.} This work was supported by the National Natural Science Foundation of China (No. 62306025, No. 92367204).
% }

% integration

{
    \small
    \bibliographystyle{ieeenat_fullname}
    \bibliography{main}
}

\clearpage
\setcounter{page}{1}
\maketitlesupplementary

\renewcommand{\thesection}{\Alph{section}}
\renewcommand {\thetable} {S\arabic{table}}
\renewcommand {\theequation} {S\arabic{equation}}
\renewcommand {\thefigure} {S\arabic{figure}}

This supplementary document is organized as follows:  1) section \ref{sec:bimodal}: more details on Bimodal Integration (BIG); 2) section \ref{sec:agq}: more quantitative studies of Adaptive Granularity Quantization (AGQ); 3) section \ref{sec:visual}: more qualitative results for instance segmentation.

\section{More Details on BIG}
\label{sec:bimodal}

\begin{figure}[h]
\begin{center}
     \includegraphics[width=1\linewidth]{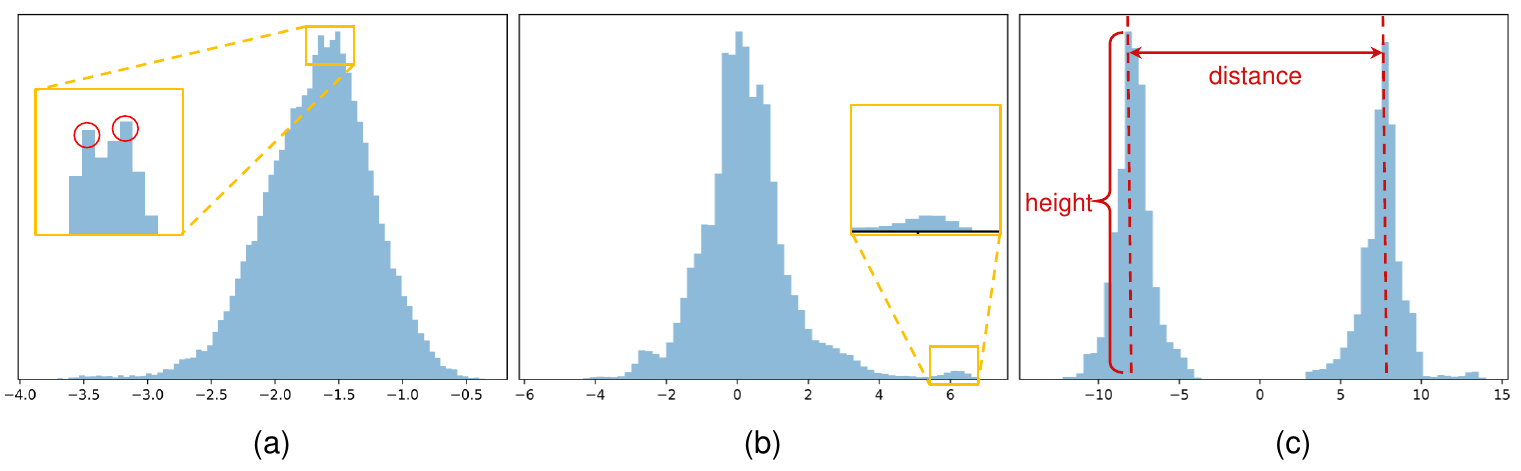}
  \caption{Three typical examples in BIG strategy.}
  \vspace{-1em}
    \label{fig:supp_discovery}
\end{center}
\end{figure}

\subsection{Bimodal Discovery}

As we mentioned in the main paper, we utilize the continuous probability density function to characterize the peaks. However, merely using the naive local maxima will induce an over-detection issue. We summarize the issue in two situations: 1) Two neighboring bumps in one peak are recognized as two peaks (Figure~\ref{fig:supp_discovery}(a)). 2) Wrongly consider the small bump as a peak (Figure~\ref{fig:supp_discovery}(b)). To address it, we impose constraints stipulating that both the peak height and the distances between two peaks must exceed a predetermined threshold in Figure~\ref{fig:supp_discovery}(c). Smaller peaks are removed first until the condition is fulfilled for all remaining peaks.

%we formulate two rules to impose constraints. Specially, we constrain the 

\begin{figure}[t]
\begin{center}
     \includegraphics[width=1\linewidth]{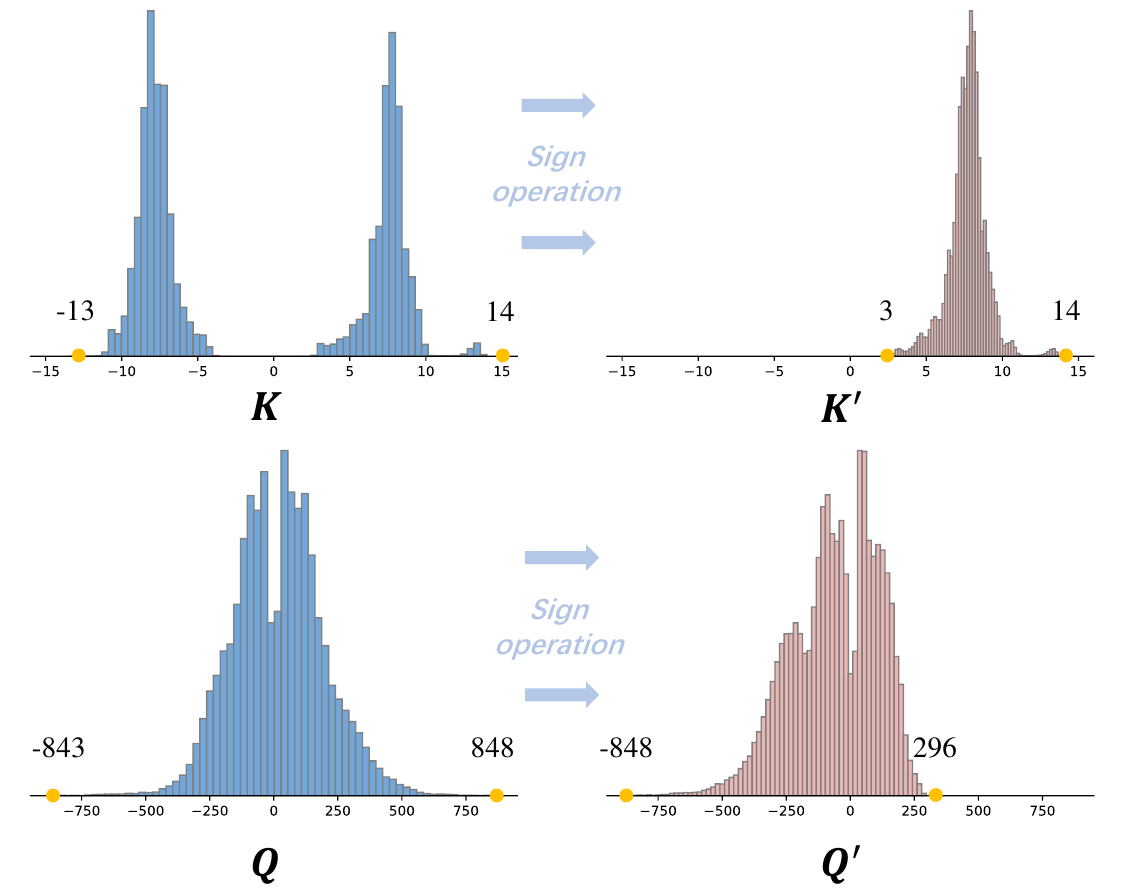}
  \caption{The distribution of query and key activations before and after BIG strategy.}
    \label{fig:supp_sign}
    \vspace{-1em}
\end{center}
\end{figure}

\subsection{Effect of Sign Operation}
To verify the effectiveness of our BIG strategy, we show the representative real distributions of query and key activations before and after sign operation. As shown in Figure~\ref{fig:supp_sign}, after sign operation, the bimodal \cls{post-Key-Linear} distribution will be transferred to a normal distribution, narrowing the range from -13$\sim$14 to 3$\sim$14 (row 1). Meanwhile, the query activations remain normal distribution invariantly, slightly reducing the range from -843$\sim$848 to -848$\sim$296 (row 2). Intuitively, our BIG is beneficial for quantization and the sign operation can be performed in advance.

\begin{figure}[t]
\begin{center}
     \includegraphics[width=1\linewidth]{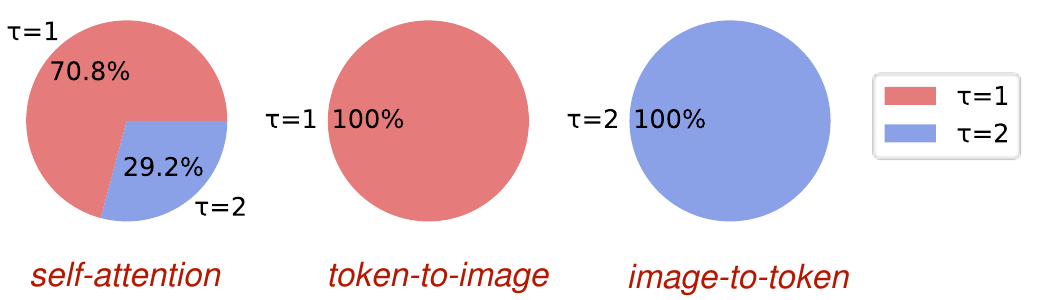}
  \caption{Pie charts depicting the optimal $\tau$ across various attention mechanisms in SAM-L.}
    \vspace{-1em}
    \label{fig:supp_pie}
    
\end{center}
\end{figure}

\begin{table}[b]
    \centering
     \resizebox{\columnwidth}{!}
     {
    \begin{tabular}{ccccccc}
    \toprule
    \textbf{Model} & \multicolumn{2}{c}{\textbf{SAM-B}}&\multicolumn{2}{c}{\textbf{SAM-L}} &\multicolumn{2}{c}{\textbf{SAM-H}}\\
    \cmidrule(lr){1-1}\cmidrule(lr){2-3}\cmidrule(lr){4-5}\cmidrule(lr){6-7}
    \textbf{\#bits} & W6A6 & W4A4 & W6A6 & W4A4& W6A6 & W4A4 \\
    \midrule
MSE$^s$   &    30.2   &   14.4   &    35.7   &    28.3  &    36.5   &    32.6   \\
\textbf{MSE$^o$}   &    \textbf{30.3}   &    \textbf{16.0}   &    \textbf{35.8}   &    \textbf{28.7}&    \textbf{36.5}   &    \textbf{33.5}   \\

      \bottomrule
    \end{tabular} 
     } % resizebox
    \vspace{-0.2mm}
    \caption{Objective test for instance segmentation. $^s$ represents quantization error for post-Softmax activations and $^o$ means quantization error for output activations of matrix multiplication.}
    \vspace{-0.8mm}
    \label{table:toy_mse}
\end{table}
\begin{figure*}[t] \centering
    \newcommand{\hwidth}{1pt}

    % \\
    % \includegraphics[width=0.24\textwidth]{figure/vis/brecq/000000579818.jpg}
    % \includegraphics[width=0.24\textwidth]{figure/vis/qdrop/000000579818.jpg}
    % \includegraphics[width=0.24\textwidth]{figure/vis/our/000000579818.jpg}
    % \includegraphics[width=0.24\textwidth]{figure/vis/fp/000000579818.jpg}
    % \\[2pt]
    
    \includegraphics[width=0.24\textwidth]{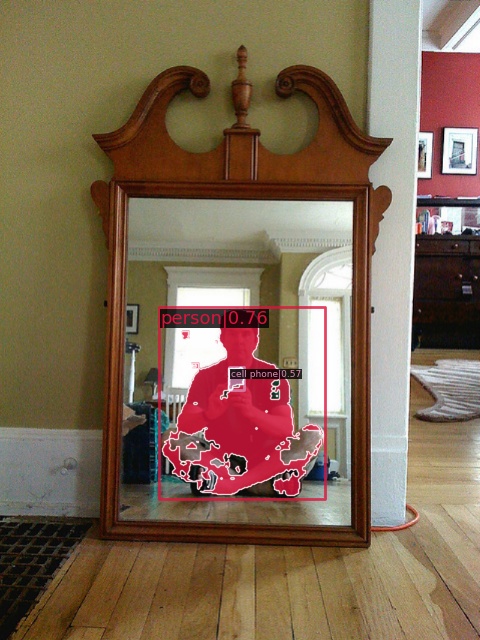}
    \includegraphics[width=0.24\textwidth]{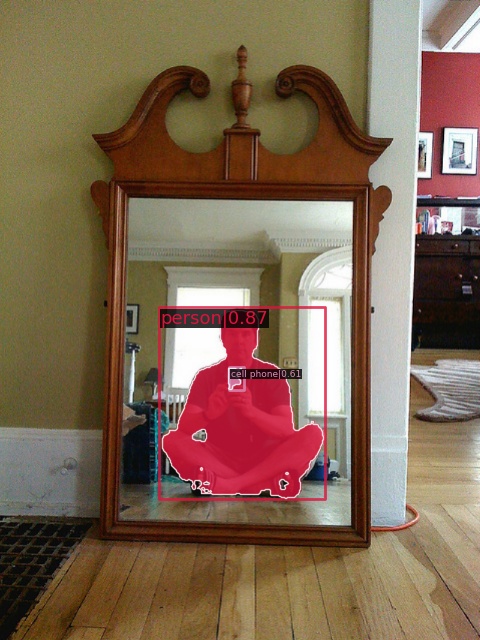}
    \includegraphics[width=0.24\textwidth]{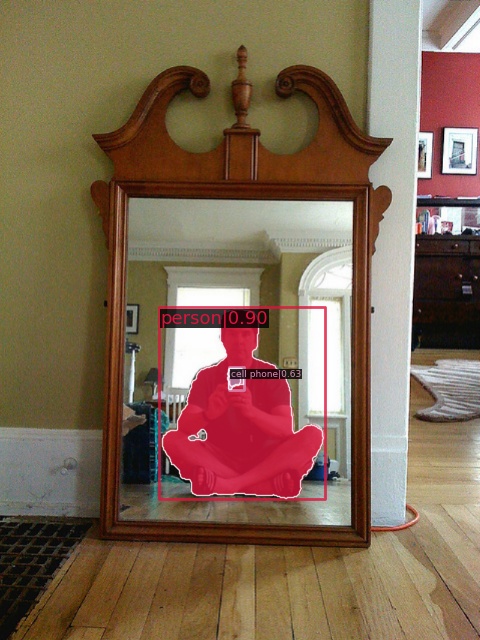}
    \includegraphics[width=0.24\textwidth]{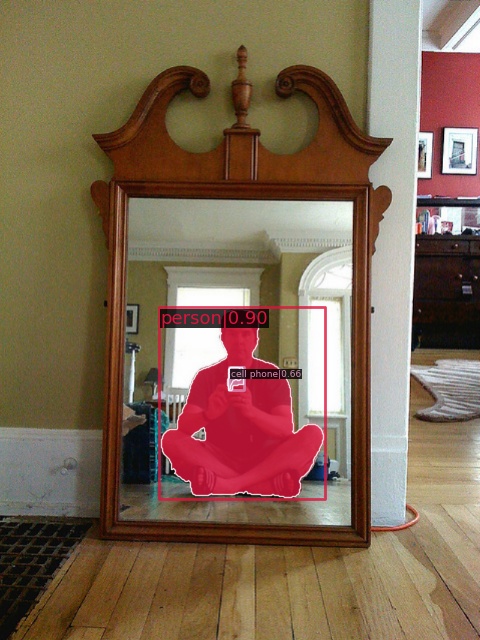}
    \\[2pt]

    % \includegraphics[width=0.24\textwidth]{figure/vis/brecq/000000577864.jpg}
    % \includegraphics[width=0.24\textwidth]{figure/vis/qdrop/000000577864.jpg}
    % \includegraphics[width=0.24\textwidth]{figure/vis/our/000000577864.jpg}
    % \includegraphics[width=0.24\textwidth]{figure/vis/fp/000000577864.jpg}
    % \\[2pt]
    \includegraphics[width=0.24\textwidth]{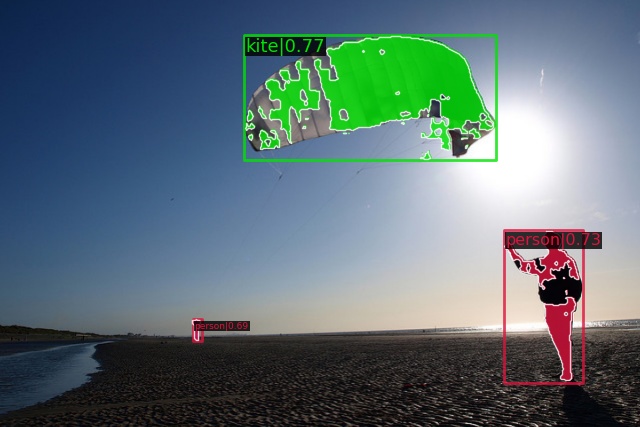}
    \includegraphics[width=0.24\textwidth]{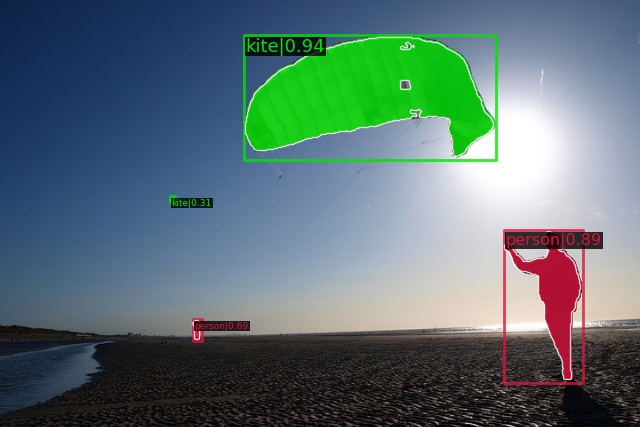}
    \includegraphics[width=0.24\textwidth]{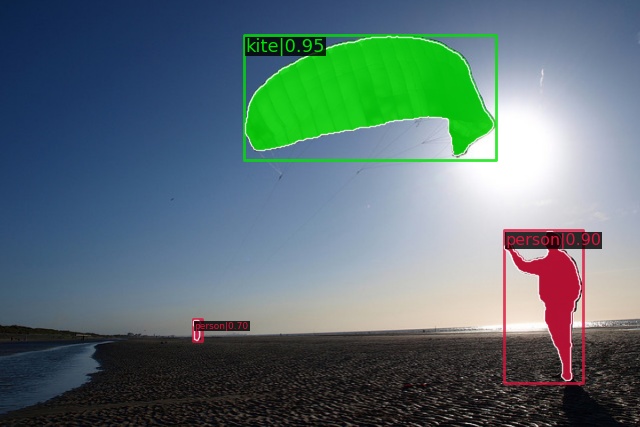}
    \includegraphics[width=0.24\textwidth]{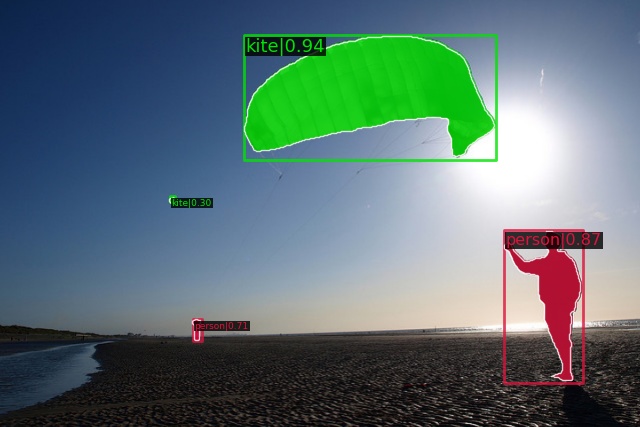}
    \\[2pt]

    \includegraphics[width=0.24\textwidth]{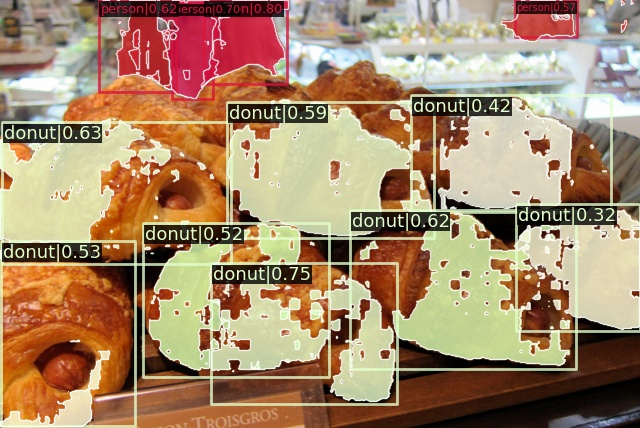}
    \includegraphics[width=0.24\textwidth]{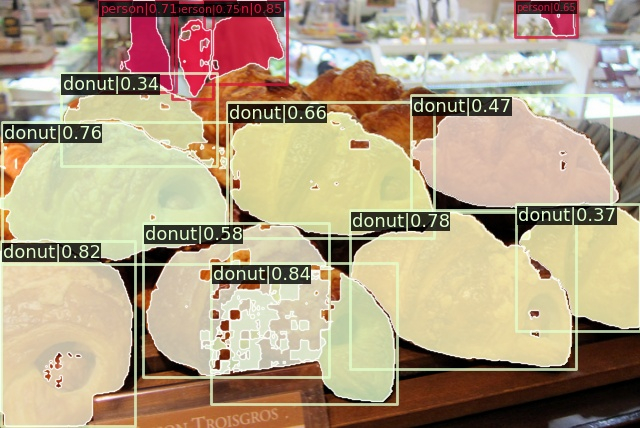}
    \includegraphics[width=0.24\textwidth]{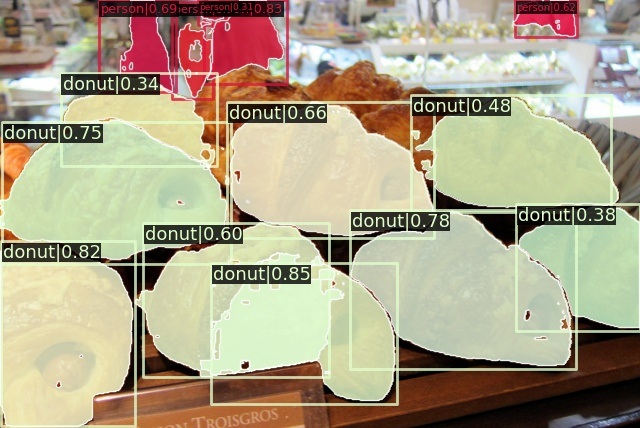}
    \includegraphics[width=0.24\textwidth]{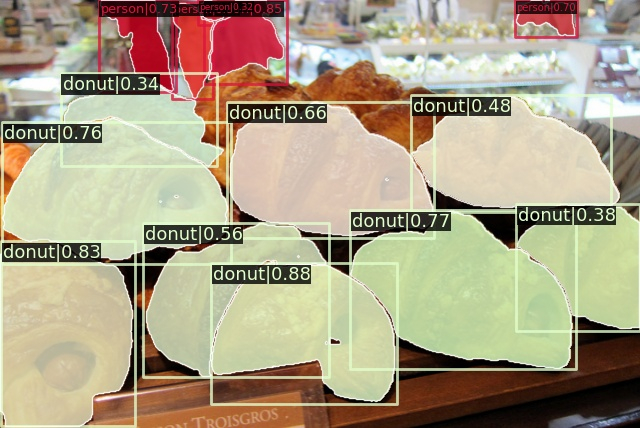}
    \\[2pt]

    \includegraphics[width=0.24\textwidth]{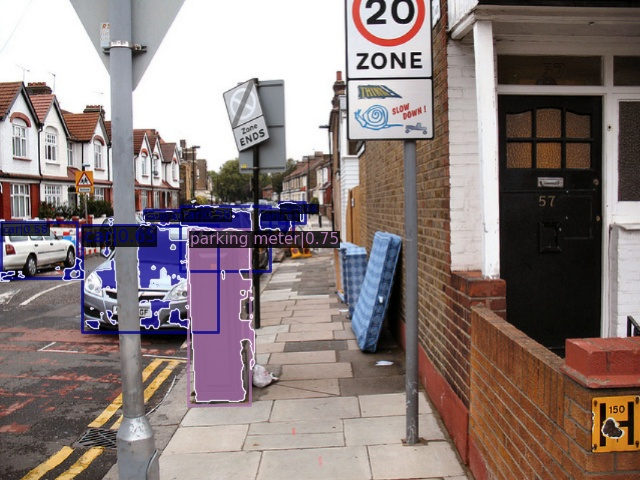}
    \includegraphics[width=0.24\textwidth]{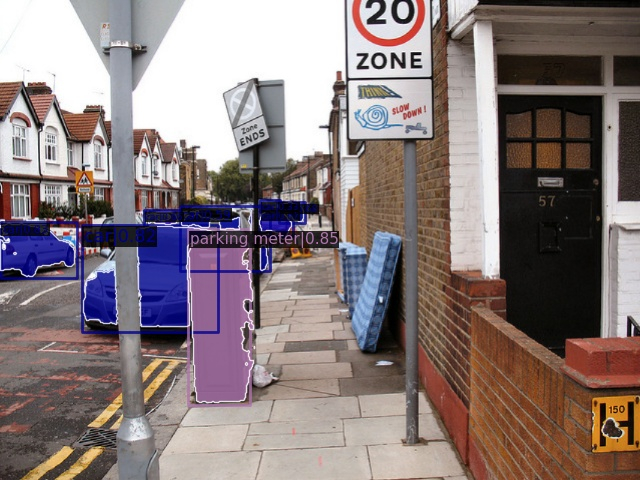}
    \includegraphics[width=0.24\textwidth]{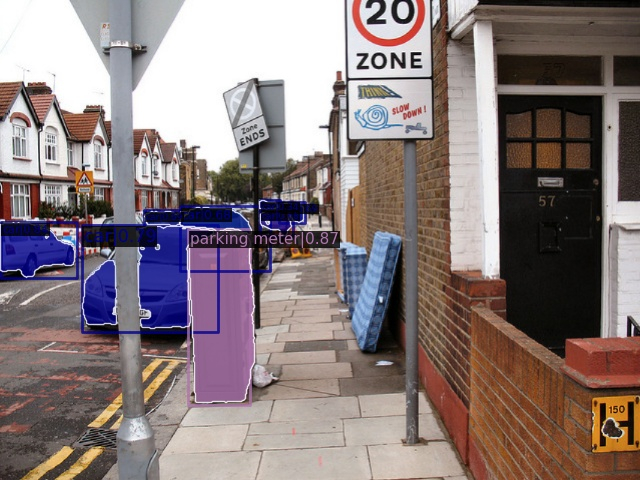}
    \includegraphics[width=0.24\textwidth]{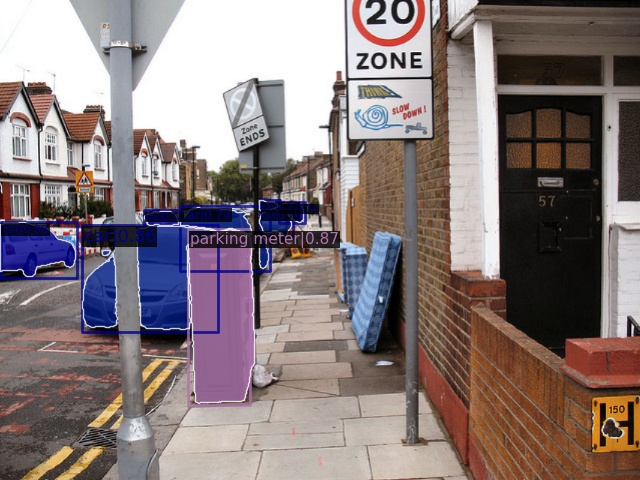}
    \\[2pt]
    
    \includegraphics[width=0.24\textwidth]{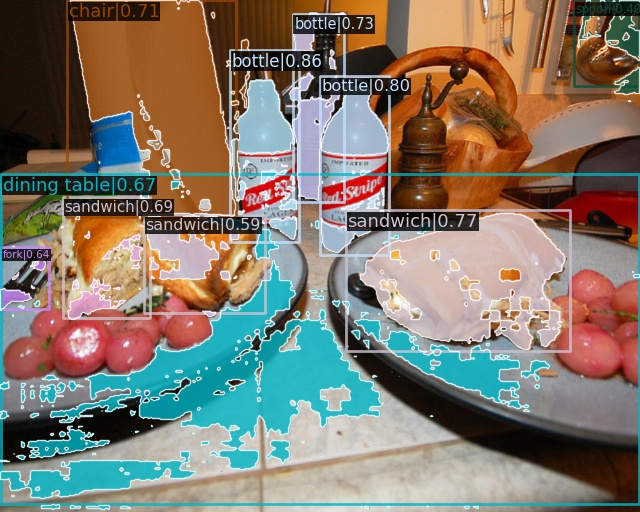}
    \includegraphics[width=0.24\textwidth]{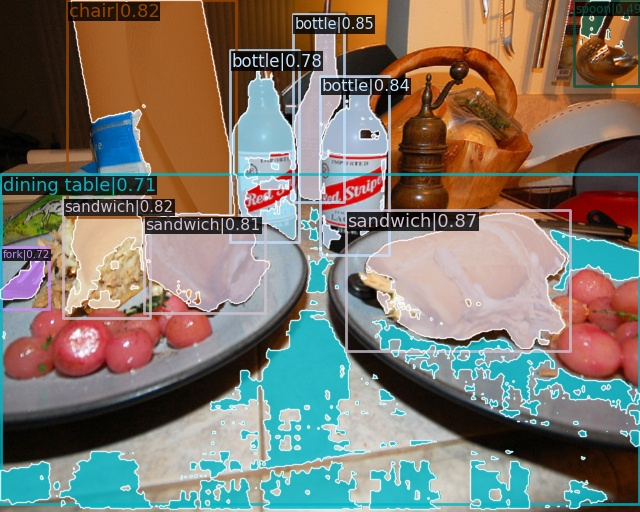}
    \includegraphics[width=0.24\textwidth]{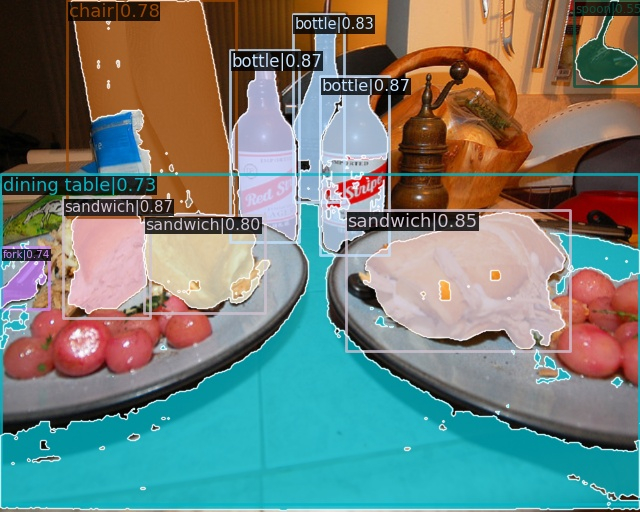}
    \includegraphics[width=0.24\textwidth]{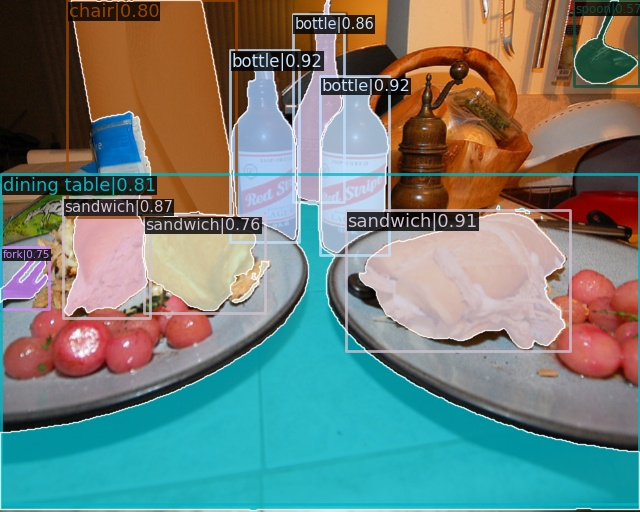}
    \\[2pt]

    \makebox[0.24\textwidth]{\small BRECQ}
    \makebox[0.24\textwidth]{\small QDrop}
    \makebox[0.24\textwidth]{\small Ours}
    \makebox[0.24\textwidth]{\small FP}

    % \makebox[0.24\textwidth]{%
    %     \includegraphics[width=0.115\textwidth]{example-image}\hspace{\hwidth}
    %     \includegraphics[width=0.115\textwidth]{example-image}} 
    % \makebox[0.24\textwidth]{%
    %     \includegraphics[width=0.115\textwidth]{example-image}\hspace{\hwidth}
    %     \includegraphics[width=0.115\textwidth]{example-image}}
    % \makebox[0.24\textwidth]{%
    %     \includegraphics[width=0.115\textwidth]{example-image}\hspace{\hwidth}
    %     \includegraphics[width=0.115\textwidth]{example-image}}
    % \makebox[0.24\textwidth]{%
    %     \includegraphics[width=0.115\textwidth]{example-image}\hspace{\hwidth}
    %     \includegraphics[width=0.115\textwidth]{example-image}}
    % \\[-0.1em]
    % \makebox[\textwidth]{\small (a) Results on data A.} 
    % \\[0.5em]

    \caption{Visualization of instance segmentation on 4-bit SAM-L.} 
    \label{fig:fig-supp-visual}
\end{figure*}

\section{Quantitative Studies of AGQ}% Adaptive Granularity Quantization
\label{sec:agq}
We complete the discussion related to the suitable granularity (optimal $\tau$) for different scenarios. As mentioned in Section 3.3, a smaller $\tau$ can better quantize lower attention scores. Conversely, with an increment in $\tau$, the higher attention scores can be quantized in a more fine-grained fashion. For simplicity, we conduct a statistical analysis of optimal $\tau$ across diverse post-Softmax distributions at W4A4. As illustrated in Figure~\ref{fig:supp_pie}, in \cls{token-to-image}, our AGQ uniformly favors $\tau$=1 because there are more low attention scores (see Figure 1 in the main paper). In \cls{image-to-token}, $\tau$=2 is prominently selected to accurately quantize more high scores. And in \cls{self-attention}, there is a coexistence of $\tau$=1 and $\tau$=2 for the combination of both high and low attention scores. Therefore, our AGQ adopts suitable granularity solutions towards the post-Softmax distribution across diverse attention mechanisms. Additionally, we compare the loss function in Eq. 14 (row 2) with local quantization errors of the attention map $\boldsymbol{A}$ (row 1). Table~\ref{table:toy_mse} indicates that Eq. 14 addresses the inconsistent issue and achieves stable performance, especially at low-bit.

% predominance of low scores
% For simplicity, we further discuss the granularity for diverse post-Softmax distributions at W4A4. As illustrated in Figure~\ref{fig:supp_pie},

% At higher bit-widths, \eg, W6A6, our AGQ tends to preserve more fine-grained higher attention scores with a larger $\tau$. When the bit-widths become lower, \eg, W4A4, AGQ will sacrifice the granularity in higher attention scores to represent lower scores, which imply the lower scores are more crucial at extremely low-bit quantization. Therefore, for simplicity, we further discuss the granularity for diverse post-Softmax distributions at W4A4. 

% As illustrated in Figure~\ref{fig:supp_pie}, our AGQ adopt the suitable granularity solutions towards the post-Softmax distribution across diverse attention mechanisms, which is consistent with our observations in the main paper: uniformly favoring $\tau$=1 in \cls{token-to-image} cross-attention as the predominance of more low scores, $\tau$=2 in \cls{image-to-token} cross-attention to preserve more high scores accurately, and a coexistence of both within self-attention.

\section{More Qualitative Results}
\label{sec:visual}
More instance segmentation results are given in Figure~\ref{fig:fig-supp-visual} produced by 4-bit BRECQ~\cite{li2021brecq}, QDrop~\cite{wei2022qdrop}, PTQ4SAM and full-precision SAM-L. Notably, our model demonstrates superior performance in terms of both completeness and clarity when compared to other methodologies. In a simple scenario with a single object, such as the \cls{person} in row 1 and the \cls{kite} in row 2, our method is capable of providing a more comprehensive description of the object boundaries, without missing any pixels. In cases where objects overlap, as observed in rows 3 and 4, our quantized model accurately distinguishes each individual object and successfully separates them from complex backgrounds. Conversely, other methods often struggle to segment occluded objects accurately, capturing unnecessary details. Particularly when recognizing background objects like the \cls{dining table}, as depicted in row 5, the results obtained from alternative approaches exhibit notable incompleteness. Conversely, our approach excels in effectively identifying the entire object, showcasing a significant advantage over other methods.

\end{document}